\documentclass[lettersize,journal]{IEEEtran}
\usepackage{amsmath,amsfonts}
\usepackage{algorithmic}
\usepackage{algorithm}
\usepackage{array}
\usepackage[caption=false,font=footnotesize,labelfont=rm,textfont=rm]{subfig}
\usepackage{textcomp}
\usepackage{stfloats}
\usepackage{url}
\usepackage{verbatim}
\usepackage{graphicx}
\usepackage{cite}
\usepackage{bbm}
\usepackage{enumitem}
\usepackage{threeparttable}
\usepackage{color}
\usepackage{colortbl}
\usepackage{xcolor}
\usepackage{multirow}
\usepackage{amssymb}
\usepackage{booktabs} 

\begin{document}

\title{R$^2$BD: A Reconstruction-Based Method for Generalizable and Efficient Detection of Fake Images}


\author{Qingyu Liu, Zhongjie Ba{$^\dagger$},~\IEEEmembership{Member,~IEEE}, Jianmin Guo, Qiu Wang, Zhibo Wang,~\IEEEmembership{Senior Member,~IEEE}, Jie Shi,~\IEEEmembership{Senior Member,~IEEE}, and Kui Ren,~\IEEEmembership{Fellow,~IEEE}
\thanks{{$^\dagger$}Corresponding author: Zhongjie Ba.}
\thanks{Qingyu Liu, Zhongjie Ba, Qiu Wang, Zhibo Wang, and Kui Ren are with the State Key Laboratory of Blockchain and Data Security, Zhejiang University, Hangzhou, Zhejiang, China (e-mail: \{qingyuliu, zhongjieba, 12321312, zhibowang, kuiren\}@zju.edu.cn).}
\thanks{Jianmin Guo and Jie Shi are with Huawei International. (e-mail: guojm17@tsinghua.org.cn, shi.jie1@huawei.com).}}




\maketitle

\begin{abstract}
Recently, reconstruction-based methods have gained attention for AIGC image detection. These methods leverage pre-trained diffusion models to reconstruct inputs and measure residuals for distinguishing real from fake images. Their key advantage lies in reducing reliance on dataset-specific artifacts and improving generalization under distribution shifts. However, they are limited by significant inefficiency due to multi-step inversion and reconstruction, and their reliance on diffusion backbones further limits generalization to other generative paradigms such as GANs.

In this paper, we propose a novel fake image detection framework, called R$^2$BD, built upon two key designs: (1) G-LDM, a unified reconstruction model that simulates the generation behaviors of VAEs, GANs, and diffusion models, thereby broadening the detection scope beyond prior diffusion-only approaches; and (2) a residual bias calculation module that distinguishes real and fake images in a single inference step, which is a significant efficiency improvement over existing methods that typically require 20$+$ steps.

Extensive experiments on the benchmark from 10 public datasets demonstrate that R$^2$BD is over 22$\times$ faster than existing reconstruction-based methods while achieving superior detection accuracy. In cross-dataset evaluations, it outperforms state-of-the-art methods by an average of 13.87\%, showing strong efficiency and generalization across diverse generative methods. The code and dataset used for evaluation are available at \url{https://github.com/QingyuLiu/RRBD}.

\end{abstract}


\section{Introduction}
\label{sec:intro}



\IEEEPARstart{W}{ith} rapid advances in deep learning, malicious attackers can now generate photorealistic facial images and videos within seconds, commonly referred to as deepfakes. Initially, deepfake techniques primarily focused on local facial manipulations generated by GANs, such as face swapping~\cite{korshunova2017fast,chen2020simswap} and expression editing~\cite{thies2016face2face,thies2019deferred}. However, the advent of diffusion models~\cite{ho2020denoising} and text-to-image (T2I) generation models~\cite{ramesh2022hierarchical,rombach2022high} has significantly amplified this threat. These advanced technologies now enable the creation of fully synthetic, high-quality facial content using nothing more than a simple text prompt. The proliferation of deepfake content poses severe risks to societal security. For example, a deepfake pornography emergency occurred in South Korea in 2024, where a huge amount of deepfake pornographic content targeting young women and even minors spread wildly on social media~\cite{Korean}, sparking widespread public concern over content safety. Therefore, developing effective methods to authenticate images and videos has become a critical concern in mitigating these potential security threats.

\begin{figure}[t]
    \centering
    \includegraphics[width=\linewidth]{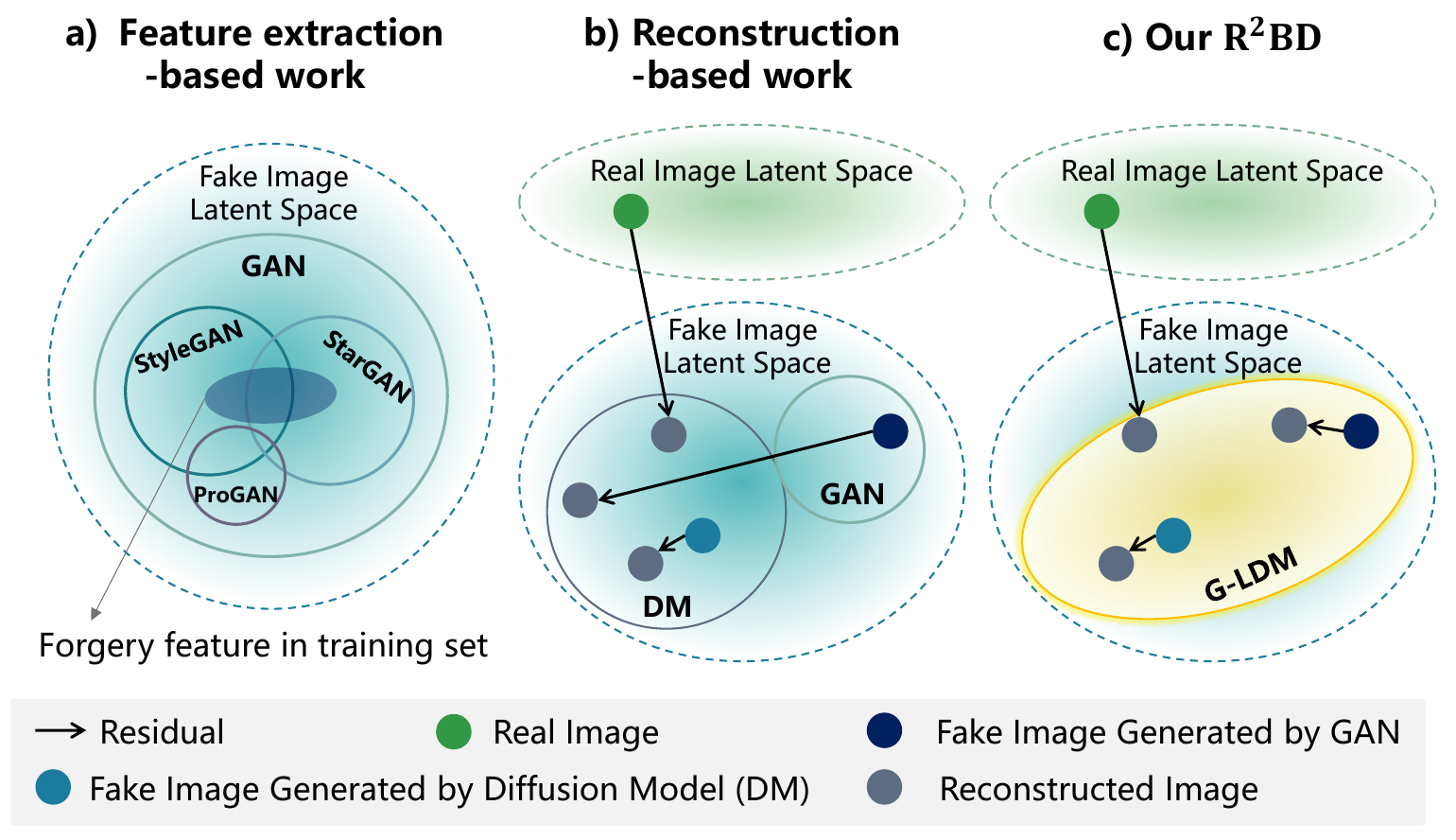}
    \caption{Illustration for observations and problems of existing works. a) Feature extraction-based works rely on the forgery features on the training dataset (e.g., StyleGAN and StarGAN), which are difficult to generalize to detecting ProGAN-based images.
b) Existing reconstruction-based methods employ diffusion models for reconstruction, which results in large residuals for both GAN-generated and real images, increasing the risk of misclassification. 
c) Our G-LDM integrates the principles of diffusion models and GANs, enabling it to reconstruct images of diverse generative paths with smaller residuals.}
    \label{fig:observation}
\end{figure}

Despite significant research efforts, existing deepfake detection methods remain highly vulnerable to generalization failures. Most prior works~\cite{rossler2019faceforensics++,shuai2023locate,ba2024exposing} rely on supervised feature learning from known forgeries, which tends to overfit to dataset-specific artifacts. As a result, these detectors often struggle when encountering forgeries generated by unseen models or novel synthesis techniques. In response, reconstruction-based approaches~\cite{wang2023dire,sha2024zerofake} have emerged as a promising alternative. By leveraging pre-trained generative models, these methods detect forgeries by measuring the difference between an image and its reconstruction, based on the empirical observation that real images are generally harder to reconstruct than fake ones. This approach reduces reliance on dataset-specific artifacts and offers improved generalization to distribution shifts.

However, reconstruction-based methods still suffer from two fundamental limitations: inefficiency and limited cross-paradigm generalization. The first limitation concerns inefficiency. These methods typically rely on pre-trained diffusion backbones and require multiple rounds of inversion and reconstruction. For instance, DIRE~\cite{wang2023dire} involves more than 20 DDIM inversion steps, resulting in an average detection time of 15.9 seconds per image, while ZeroFake~\cite{sha2024zerofake} requires up to 999 steps, with an average of 82.33 seconds per image. Such heavy computational overhead severely restricts their practicality for real-time detection or large-scale deployment.

The second limitation lies in cross-paradigm generalization. Existing reconstruction-based detectors perform well only when images are generated under the diffusion paradigm. When extended to non-diffusion paradigms such as GANs, their generalization performance degrades significantly. This weakness stems from the fact that diffusion-based reconstruction models are intrinsically constrained to approximating diffusion trajectories. Since different generative paradigms (e.g., VAEs, GANs, diffusion) induce distinct statistical priors and reconstruction behaviors, directly applying a diffusion-only model to reconstruct non-diffusion forgeries is inherently mismatched and undermines generalization.

In this paper, we propose \textbf{R}econstruction-based \textbf{R}esidual \textbf{B}ias for fake image \textbf{D}etection (R$^2$BD), a framework designed to address both inefficiency and cross-paradigm generalization challenges. It consists of three key components: (1) a unified \textbf{G}AN-\textbf{L}atent \textbf{D}iffusion \textbf{M}odel (G-LDM) that simulates mainstream synthesis mechanisms, (2) a residual bias calculation technique that extracts discriminative signals in a single inference step, and (3) a lightweight two-stream classifier that fuses RGB and latent residual bias for final prediction.

To mitigate the cross-paradigm generalization challenge, we first design G-LDM, a unified reconstruction model that integrates VAE-, GAN-, and diffusion-based generation behaviors. Built upon the Stable Diffusion architecture, G-LDM retains the latent-space denoising formulation while incorporating adversarial training. We view the latent U-Net as a conditional generator and pair it with a discriminator to guide GAN-like generation trajectories. This enables G-LDM to reconstruct forgeries generated by different paradigms with low residuals, while real images consistently yield larger ones (Fig.~\ref{fig:observation}).

To tackle the efficiency issue, we design a residual bias calculation technique and derive a theoretically grounded signal beyond raw reconstruction residuals. Specifically, previous works relied on the empirical observation that real images are harder to reconstruct, but lacked a principled explanation. We instead offer a theoretical interpretation: fake images, being sampled from distributions similar to the reconstruction model, tend to exhibit residuals close to the model’s inherent reconstruction baseline. In contrast, real images contain domain-specific traces not captured by the model, resulting in residuals that deviate more from this baseline. We define residual bias as the difference between the measured residual and the model’s theoretical reconstruction residual, and observe that even with a single-step reconstruction, residual bias produces clear cluster-level separation between real and fake samples.


Finally, we employ a two-stream classifier to jointly analyze residual bias in both RGB and latent spaces. Since the theoretical and measured residuals are defined in different feature spaces, their discrepancies cannot be directly subtracted. Therefore, we compute residual bias separately in each space, where the RGB stream reflects pixel-level deviations and the latent stream encodes high-level reconstruction inconsistencies. The two-stream classifier then fuses these complementary signals to provide final real–fake predictions.

We evaluate R$^2$BD on the benchmark constructed from 8 open-source datasets and two commercial AIGC APIs (Midjourney and DALL·E), covering three major generative paradigms: GANs, pixel-based diffusion, and latent diffusion. The evaluation includes both in-dataset tests and cross-dataset tests on 13 unseen forgery methods to assess generalization across and within paradigms. Extensive experiments demonstrate that R$^2$BD achieves superior performance under both single-paradigm and hybrid training settings. In particular, in cross-dataset evaluations, R$^2$BD outperforms existing methods by an average of 13.87\% in hybrid training, with a 6.72\% gain specifically for GAN-based AIGC images. Moreover, R$^2$BD requires only 0.706 seconds per image, which is over 22$\times$ faster than related reconstruction-based approaches and comparable to a binary classifier. We further conduct robustness evaluations under various perturbations and ablation studies to validate the effectiveness of each component. In summary, our contributions are as follows:
\begin{itemize}[leftmargin=*]

\item We propose R$^2$BD, a new reconstruction-based detection framework for AIGC images that improves efficiency and cross-paradigm generalizability.


\item To tackle the efficiency issue of existing reconstruction-based methods that require more than dozens of inversion steps, we propose a novel one-step residual bias calculation technique. It achieves efficiency comparable to binary classifiers and runs over 22$\times$ faster (15.9s $\rightarrow$ 0.706s) than existing methods.

\item To enhance cross-paradigm generalization, we introduce G-LDM, a reconstruction model that unifies the generative behaviors of VAEs, GANs, and diffusion models within a single latent diffusion model. This enables our R$^2$BD framework to overcome the limitation of diffusion-only detection in existing methods.

\item Extensive evaluations demonstrate that R$^2$BD consistently outperforms existing methods in detecting images generated by different generative paradigms. 
In hybrid training settings, it achieves an average accuracy gain of 13.87\% over state-of-the-art detectors.
\end{itemize}    
\section{Related Work}  
\label{sec:relwork}
In this section, we discuss existing AIGC techniques and fake image detection methods.

\subsection{AIGC Image Generation}

Automatic image generation has witnessed significant progress with the rise of three dominant generative paradigms: VAEs, GANs, and Diffusion Models.

VAE-based methods model the data distribution by learning a latent representation and a decoder that reconstructs input images. While early works such as VAE~\cite{kingma2013auto} and its extensions (e.g., VQ-VAE~\cite{van2017neural}) offered stable training and interpretability, the generated images often lacked sharpness and fine details due to blurry reconstructions caused by the pixel-wise loss.

GAN-based methods, introduced by Goodfellow et al.\cite{goodfellow2020generative}, formulate image synthesis as a minimax game between a generator and a discriminator. They have enabled high-fidelity and high-resolution image synthesis, especially with architectural advances like StyleGAN2\cite{karras2020analyzing}, StyleGAN3~\cite{karras2021alias}, and Projected GAN~\cite{Sauer2021NEURIPS}. These models can produce photo-realistic faces and fine-grained image edits but suffer from training instability and vulnerability to mode collapse.

Diffusion-based models represent a more recent and powerful class of generative techniques that model the data generation process as the reverse of a gradual noising process. Pioneered by DDPM~\cite{ho2020denoising}, subsequent improvements include DDIM~\cite{song2021denoising}, ADM~\cite{dhariwal2021diffusion}, and latent diffusion models (LDMs) such as Stable Diffusion~\cite{rombach2022high}. These models have demonstrated impressive scalability and controllability, especially in conditional generation and high-resolution synthesis. Notably, diffusion models are adopted in both pixel space (e.g., ADM, DDPM) and latent space (e.g., Stable Diffusion, DALL$\cdot$E 2~\cite{ramesh2022hierarchical}, and Midjourney~\cite{midjouney}).

\vspace{-2mm}
\subsection{AIGC Image Detection}
\vspace{-1mm}

The increasing ubiquity of AIGC, particularly synthetic images produced by GANs and diffusion models, has created a pressing need for reliable and generalizable fake image detection. Over time, detection techniques have followed a clear evolutionary path, progressing from early feature-based classifiers to more recent reconstruction-based and vision-language-based approaches. This trajectory reflects a growing trend toward incorporating knowledge of the generative process into the detection pipeline.

Early detection methods approached fake image identification as a binary classification task, aiming to learn discriminative features directly from visual inputs. These approaches typically relied on improved network architectures~\cite{rossler2019faceforensics++,shuai2023locate,zhao2021multi} or refined loss functions~\cite{ba2024exposing} to capture subtle artifacts left by generative models. In some cases, frequency-based representations~\cite{masi2020two,luo2021generalizing,liu2021spatial,qian2020thinking} were also incorporated to enhance sensitivity to periodic or high-frequency artifacts often introduced by upsampling operations in generative models. While effective when the test distribution matches the training data, such methods often exhibit poor generalization across different generative paradigms. As noted previously, detectors trained on a specific type of generator (e.g., GANs) often fail to detect forgeries produced by other paradigms (e.g., diffusion models), largely due to the absence of shared forensic cues. In addition, these methods are predominantly data-driven and lack interpretability or meaningful forensic insight.

To address the limitations of generalization and transparency, recent research has explored the use of large-scale pre-trained models. One notable direction involves vision-language model (VLM)-based detection~\cite{zhang2024common,chen2024textit}, which reframes the task as a multimodal reasoning problem rather than a simple classification exercise. For example, ~\cite{zhang2024common} utilizes vision-language transformers to generate natural language explanations grounded in visual evidence, enabling human-aligned interpretations of deepfake artifacts. While these methods enhance interpretability through semantic reasoning, they fall short in detecting fine-grained manipulation artifacts and generalizing across heterogeneous generative models. As such, VLM-based methods are typically used as auxiliary tools rather than standalone solutions for AIGC detection.

Another prominent line of work~\cite{wang2023dire,sha2024zerofake,chen2024drct,luo2024lare} focuses on reconstruction-based detection. This paradigm is grounded in the empirical observation that real images are generally more difficult to reconstruct than fake ones. Most methods employ a pre-trained diffusion model to reconstruct the input and then compute residuals for classification. For instance, DIRE~\cite{wang2023dire} adopts ADM as the reconstruction backbone and performs detection using pixel-space residuals. ZeroFake~\cite{sha2024zerofake} instead builds on Stable Diffusion, incorporating text prompts during reconstruction, and reports that fake images exhibit greater robustness to adversarial perturbations than real ones. Compared with feature-based methods, reconstruction-based approaches typically achieve stronger generalization under distribution shifts, as they leverage generative priors. However, their dependence on diffusion models trained within a specific generative paradigm (e.g., ADM or Stable Diffusion) restricts adaptability to other generator families such as GANs. To address this issue, recent work~\cite{jiang2025model} reconstructs images using multiple pre-trained generative models to capture paradigm-specific discrepancies. In contrast, our framework accomplishes this goal through a unified reconstruction model, thereby eliminating the need for multiple independent generators. Nonetheless, such multi-reconstruction strategies still introduce substantial computational overhead, which limits their practicality for real-world deployment.

More recently, LaRE$^2$~\cite{luo2024lare} proposed a hybrid detection framework that integrates both feature-based and reconstruction-based cues. Instead of treating the reconstruction residual as an independent input, LaRE$^2$ uses it as a guiding signal to enhance feature extraction from the original image. This design improves efficiency compared with previous reconstruction-based methods. However, it still fundamentally depends on the presence of forgery artifacts in the original image and thus inherits the generalization limitations of feature-based approaches under distribution shifts. In contrast, our method adopts a purely reconstruction-based framework that bypasses the original image entirely. Inspired by the intrinsic difference in reconstruction behavior between real and fake images, we introduce the concept of residual bias, which refers to the discrepancy between measured and theoretical residuals and can be computed in a single reconstruction step. This formulation allows us to directly distinguish real and fake images with minimal computational overhead. As a result, our method achieves both high efficiency and strong generalization across diverse generative paradigms. Moreover, it can be flexibly extended to a dual-branch framework, where residual bias and original image features are jointly analyzed to enhance detection performance through complementary signals.
\section{Threat Model}
\label{sec:threat_model}
In this section, we introduce the application scenarios, threat model, and design goals of the proposed R$^2$BD.  

\subsection{Adversary Model}

\textbf{Adversary Motivation.}
The goal of face manipulators (i.e., adversaries) is to generate or manipulate the target person’s face without the owner's authorization, in order to achieve their malicious objectives. Specifically, there are three common motivations in real-world scenarios:

{1) Financial Fraud:} Adversaries use deepfake technology to fabricate the victim's voice or image in order to commit fraud or economic scams. A recent example is the Hong Kong deepfake scam case in Feb. 2024, where attackers used simulated video and audio of company executives to deceive financial personnel into transferring 200 million Hong Kong dollars.

{2) Spreading False Imagery:} Malicious actors may leverage AIGC technology to insert the victim into inappropriate or violent scenes, thereby disseminating misleading information. For example, in the "Deepfake" group incident in South Korea in August 2024, criminals utilized deepfake technology to create pornographic materials involving hundreds of women and disseminated them on public social media platforms.

{3) Bypassing Facial Authorization:} The use of deepfake technology has the potential to circumvent the facial verification process, which is becoming increasingly prevalent in security-critical applications. For instance, criminals could exploit this technology to bypass a bank's facial authentication system, thereby facilitating a fraudulent money transfer.

\textbf{Adversary Assumption.}
We assume the adversaries can:
1) Utilize any state-of-the-art generation techniques or third-party AIGC APIs to create or modify facial images, such as GANs, Diffusion Models, T2I services, etc;
2) Deploy image degradation techniques, including compression, noise, blur, and color manipulation, to simulate low real-world image quality and reduce the effectiveness of deepfake detection methods.

\subsection{System Model}
\label{sec:design_goal}
In this paper, we focus on face-related scenarios that pose greater security threats. Specifically, a user (also called a defender) can use the proposed R$^2$BD to distinguish real and fake images generated by various forgery methods. We assume that the defender has little knowledge of the specific forgery model used by adversaries. Our design goals are:
\begin{itemize}[leftmargin=*,itemsep=2pt,topsep=0pt,parsep=0pt]
    \item \textbf{Effectiveness.} The primary goal of our detector is to effectively differentiate between generated and real facial images, including those produced using GANs and diffusion models.
    \item \textbf{Generalization.} The detector should maintain strong generalization performance against various unknown forgery methods including GAN, diffusion model, and T2I generation model.
    \item \textbf{Efficiency.} As mentioned above, existing reconstruction-based detection methods suffer from heavy reconstruction inference steps, making them impractical. Our goal is to propose a detection method with fewer inference steps to achieve highly efficient detection.
\end{itemize}

\section{Methodology}
\label{sec:methodology}
In this section, we start by introducing the overview of proposed framework R$^2$BD, and then present two main modules one by one, including G-LDM model and residual bias calculation module.
\subsection{Overview}

\textbf{Training.} In this paper, we design and implement a generalizable and efficient fake image detection framework R$^2$BD. Its training process consists of three stages: reconstruction model training, residual bias calculation, and detector training.

1) Training reconstruction model: 
We introduce a novel reconstruction model named G-LDM, a GAN-enhanced variant of the Latent Diffusion Model (LDM). In G-LDM, the original UNet component of LDM is repurposed as a generator within a GAN architecture. A discriminator is simultaneously trained to distinguish real latent features from those reconstructed by the generator, while the generator learns to minimize both the standard diffusion reconstruction loss and an adversarial loss. By jointly optimizing diffusion and adversarial objectives, G-LDM integrates the training principles of VAEs, GANs, and diffusion models into a unified framework. This hybrid formulation enables G-LDM to simulate diverse generative processes and approximate the underlying generation pathways of various synthetic image sources. Consequently, G-LDM achieves consistently lower reconstruction errors across different generative models, providing a robust foundation for downstream residual analysis and classification. The training process ultimately produces a pre-trained G-LDM model, which serves as the backbone of the R$^2$BD framework.

2) Calculating residual bias:
We introduce a new residual bias calculation module to preprocess data for detector training. This module aims to quantify the bias between the theoretical residual and the actual measured residual during the reconstruction process. The theoretical residual is analytically derived from the DDIM inversion and reconstruction equations in the latent space, while the measured residual is defined as the pixel-wise difference between the original image and its reconstruction in RGB space. We observe that the discrepancy between the theoretical and measured residuals (i.e., residual bias) is generally larger for real images than for fake ones. Based on that, we compute residual bias using the pre-trained G-LDM model described above. Since the theoretical and measured residuals lie in different representational domains, we adopt a two-stream strategy: residual bias is calculated separately in both the latent space and the RGB space, and both are used as complementary inputs for detector training. This preserves critical information at both semantic and perceptual levels, enabling the classifier to more effectively learn the discriminative residual patterns between real and fake images. Notably, the entire residual bias extraction process requires only one step of DDIM inversion and reconstruction processes through G-LDM, making it computationally efficient.

\begin{figure*}[t]
    \centering
    \includegraphics[width=0.9\linewidth]{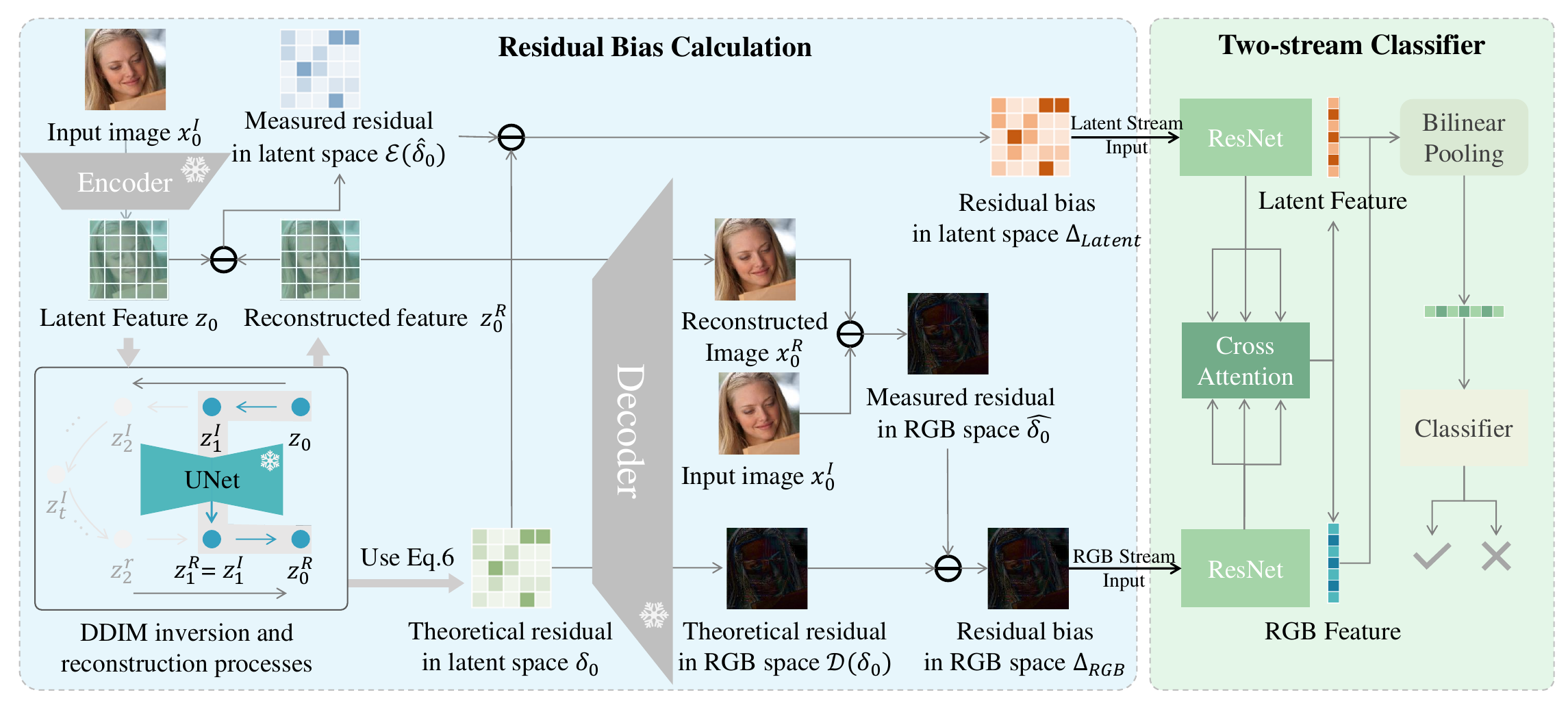}
    \caption{Overview of the R$^2$BD inference pipeline. Given an input image, we perform one-step DDIM inversion and reconstruction using the pre-trained G-LDM to compute residual bias in both latent and RGB spaces. The two residual bias features are then fed into a two-stream classifier to predict whether the image is real or fake.}
    \label{fig:inference}
\end{figure*}

3) Training detector: 
We design a lightweight two-stream classification network tailored for fake image detection, as illustrated in Fig.~\ref{fig:inference}. The network takes as input the residual bias in RGB space and the residual bias in latent space, which are independently processed through two feature extractors. The resulting feature embeddings are fused at an intermediate stage via cross-attention layers, enabling the model to capture interactions between perceptual and semantic-level reconstruction discrepancies. The fused features are then passed through a classifier to produce the final prediction.

\textbf{Detection.} At inference time, R$^2$BD operates in two stages: residual bias computation and fake image classification. As shown in Fig.~\ref{fig:inference}, given a suspicious image, we first use the pre-trained G-LDM to compute the residual biases in both latent space and RGB space through DDIM inversion and reconstruction at one step. These are then fed into the trained two-stream detector to predict whether the input image is real or fake.

\vspace{-3mm}
\subsection{G-LDM Model}
\vspace{-1mm}
\begin{figure}[t]
    \centering
    \includegraphics[width=\linewidth]{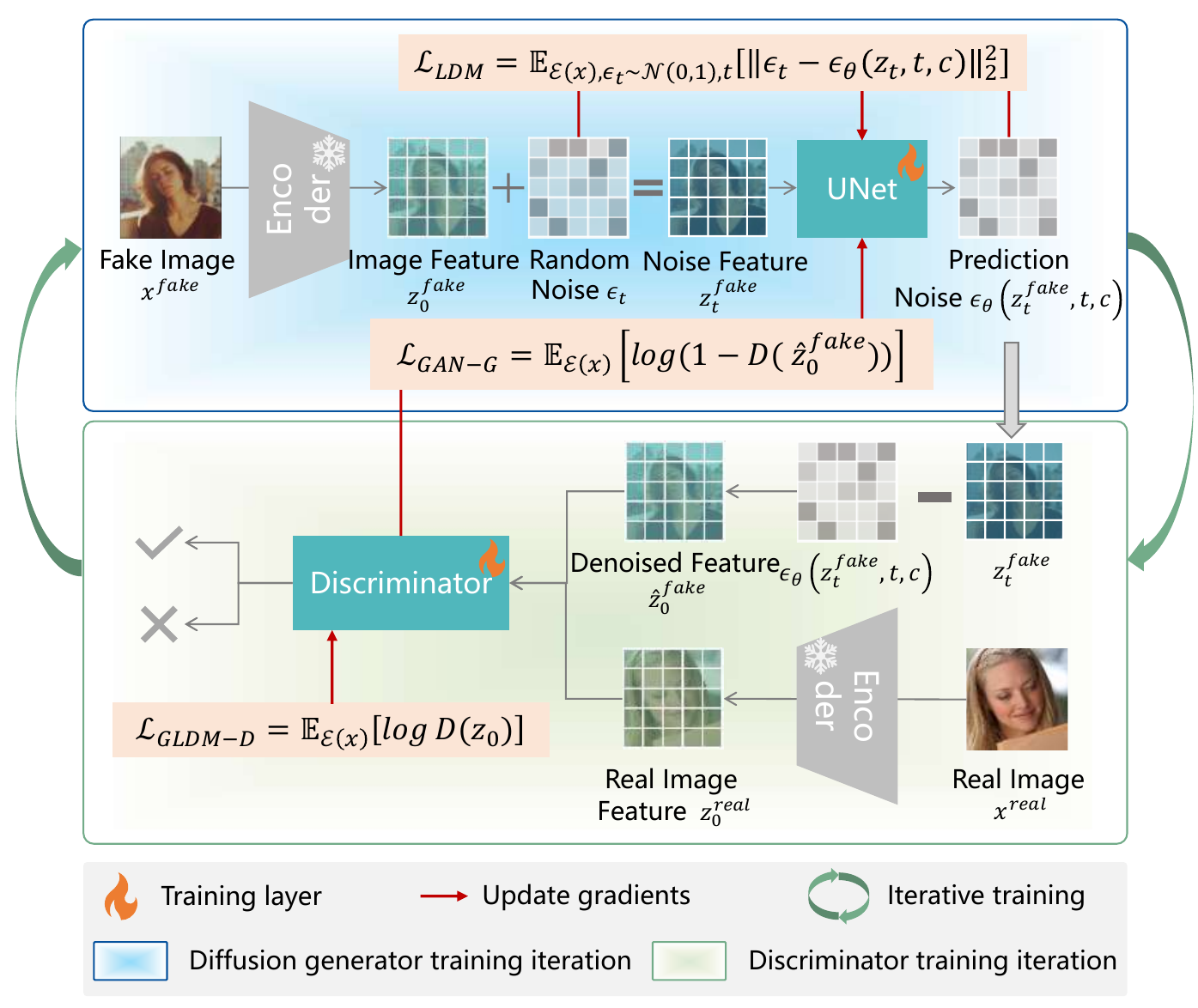}
    \caption{The training process of G-LDM model. Diffusion generator and discriminator training iterative alternating training.}
    \label{fig:G-LDM}
\end{figure}

Existing reconstruction-based AIGC detection methods leverage the strong reconstruction capability of diffusion models to recover the generation paths of suspicious images. The core observation is that the generative process of fake images can be more accurately reproduced by the generative model with smaller residuals, whereas real images typically result in larger reconstruction errors. However, most existing approaches primarily rely on pre-trained diffusion models (e.g., ADM and Stable Diffusion) to approximate the generation path, which introduces a key limitation. Specifically, in practice, the majority of AIGC methods fall into three major families: GAN-based, diffusion-based, and VAE-based generators. When a reconstruction model is trained solely with diffusion-based objectives, it lacks the flexibility to effectively reconstruct fake images produced by non-diffusion models such as GANs. As a result, the reconstruction residuals for GAN-generated images also remain large, making it difficult to distinguish them from real images. Consequently, the performance of such detection methods degrades in cross-generator scenarios.

To overcome this limitation, we propose G-LDM, a reconstruction model designed to enhance compatibility with different types of generative processes. G-LDM preserves the latent-space denoising structure of LDMs while introducing a hybrid training strategy that integrates both diffusion-based reconstruction and adversarial learning objectives, thereby improving its reconstruction fidelity across a wider range of fake image types. 

As shown in Fig.~\ref{fig:G-LDM}, the training process of our G-LDM involves alternating iterations between the diffusion generator and the discriminator. Given a fake image $x_{fake}$, we use a pre-trained encoder $\mathcal{E}$ to map $x_{fake}$ into the low-dimensional latent space, obtaining the latent representation $z_0^{fake}=\mathcal{E}\left(x_{fake}\right)$. At a randomly chosen time step $t \in (0, 1000)$, a gaussian noise $\epsilon_t$ is randomly added to $z_0^{fake}$, resulting in the noisy feature $z_t^{fake}$. Given $z_t^{fake}$, the UNet is then used to predict the noise at step $t$, resulting in the predicted noise $\epsilon_{\theta}( z_t^{fake}, t, c)$, where $c$ is embedding of text condition. This allows us to compute the difference between the added noise and the predicted noise as the diffusion loss $\mathcal{L}_{LDM}$:
\begin{equation}
\begin{aligned}
\mathcal{L}_{L D M}=\mathbb{E}_{\mathcal{E}(x), \epsilon_t \sim \mathcal{N}(0,1), t}\left[\left\|\epsilon_t-\epsilon_\theta\left(z_t, t,c\right)\right\|_2^2\right].
\end{aligned}
\label{eq:LDM_loss}
\end{equation}
Then, by subtracting the predicted noise $\epsilon_{\theta}( z_t^{fake}, t,c)$ from the noise feature $z_t^{fake}$, we can obtain the denoised feature of the fake image $\hat{z}_0^{fake}$. We then feed it into the discriminator for the authenticity judgement. We treat the UNet in the diffusion process as the generator in the GAN, so the denoised feature is considered as the generated image feature. Therefore, the goal of the diffusion generator is to fool the discriminator $D$, resulting in the generator loss $\mathcal{L}_{GAN-G}$:
\begin{equation}
\begin{aligned}
\mathcal{L}_{G A N-G}=\mathbb{E}_{\mathcal{E}(x)}\left[\log \left(1-D\left(\hat{z}_0^{fake}\right)\right)\right],
\end{aligned}
\label{eq:GAN-G_loss}
\end{equation}
where $D(\hat{z}_0^{fake})$ represents the discriminator's prediction of the authenticity of the denoised feature. Therefore, the total loss function of the diffusion generator iteration consists of the sum of the diffusion loss (Eq.~\ref{eq:LDM_loss}) and the generator loss (Eq.~\ref{eq:GAN-G_loss}), expressed as $\mathcal{L}_{GLDM-G} = \mathcal{L}_{LDM}+\mathcal{L}_{GLDM-G}$. Here, $\mathcal{L}_{GLDM-G}$ only updates the parameters of the UNet during the training process of the diffusion generator iteration, while the other parameters are frozen. Additionally, we only use fake images to train the UNet. This is because if real images were used to train the reconstruction model, there would be a smaller residual bias for real data in the training set, which could negatively affect the performance of the subsequent detector when trained on real datasets.

For the discriminator training iteration, our goal is to train a discriminator that can distinguish between true and false. Since the diffusion generator iteration above only provides negative samples for the discriminator, we additionally include real images to train the discriminator. Therefore, the loss for the discriminator training iteration is expressed as $\mathcal{L}_{G L D M-D}=\mathbb{E}_{\mathcal{E}(x)}\left[\log D\left(z_0\right)\right]$. Here, $\mathcal{L}_{G L D M-D}$ only updates the parameters of the discriminator. 

In summary, the training of G-LDM alternates between the diffusion generator iteration and the discriminator training iteration. To ensure training stability, we first train 300 diffusion generator iterations, followed by five discriminator training iterations, and then repeat this alternating process. When the training of G-LDM, we use Stable Diffusion V2.1 as the pre-trained model. Additionally, we use Llava~\cite{liu2023llava} to generate corresponding image captions for the fake images, forming image-text pairs for training G-LDM.

\vspace{-3mm}
\subsection{Residual Bias Calculation}
\vspace{-1mm}


Reconstruction-based AIGC detection has gained traction due to a widely observed empirical phenomenon: real images are significantly more difficult to reconstruct than fake ones. In this work, we propose a theoretical framework to explain and formalize this observation. We argue that authentic images contain domain-specific real-world clues that are fundamentally unlearnable by generative models and therefore cannot be faithfully reconstructed. In contrast, such clues are absent in synthetic content, meaning that AI-generated images produced from a similar model distribution can be reconstructed with only minor residuals.

This observation raises a critical question: how can we effectively extract such domain-specific traces to distinguish real from fake images? To address this challenge, we analyze the diffusion reconstruction pipeline to theoretically model and quantify the reconstruction behavior of real versus fake images.

\textbf{Preliminary.} We begin with vanilla diffusion models, where the generation process is conducted directly in the pixel space. Starting from Gaussian noise $x_T$, a learned noise predictor $\epsilon_\theta$ iteratively denoises the input to produce the final image $x_0$. The DDIM~\cite{song2021denoising} deterministic sampling process at step $t$ is defined as:
\begin{equation}
\begin{aligned}
x_{t-1}=\frac{1}{\sqrt{\alpha_t}}x_t+\left(\sqrt{1-\bar{\alpha}_{t-1}}-\frac{1}{\sqrt{\alpha_t}}\sqrt{1-\bar{\alpha}_t}\right)\epsilon_\theta\left(x_{t}, t\right).
\end{aligned}
\label{eq:ddim}
\end{equation}
where $\alpha_t$ is the noise scaling factor, and $\bar{\alpha}_t = \prod_{i=1}^{t} \alpha_i$. The corresponding inversion process attempts to reverse this trajectory from $x_0$ back to $x_T$, approximated as:
\begin{equation}
\begin{aligned}
x_t^I=\sqrt{\alpha_t}x_{t-1}^I+ \left(\sqrt{1-\bar{\alpha}_t}-\sqrt{\alpha_t-\bar{\alpha}_t}\right)\epsilon_\theta\left(x_{t-1}^I, t\right).
\end{aligned}
\label{eq:ddim_inversion}
\end{equation}
However, the approximation in inversion (e.g., using $x_{t-1}^I$ instead of the true $x_t^I$) introduces a theoretical residual across steps. We leverage this theoretical formulation to isolate a more meaningful detection feature.

\begin{figure}[t]
\centering
\subfloat[Measured residual $\hat{\delta}_0$]{\includegraphics[width=0.75\linewidth]{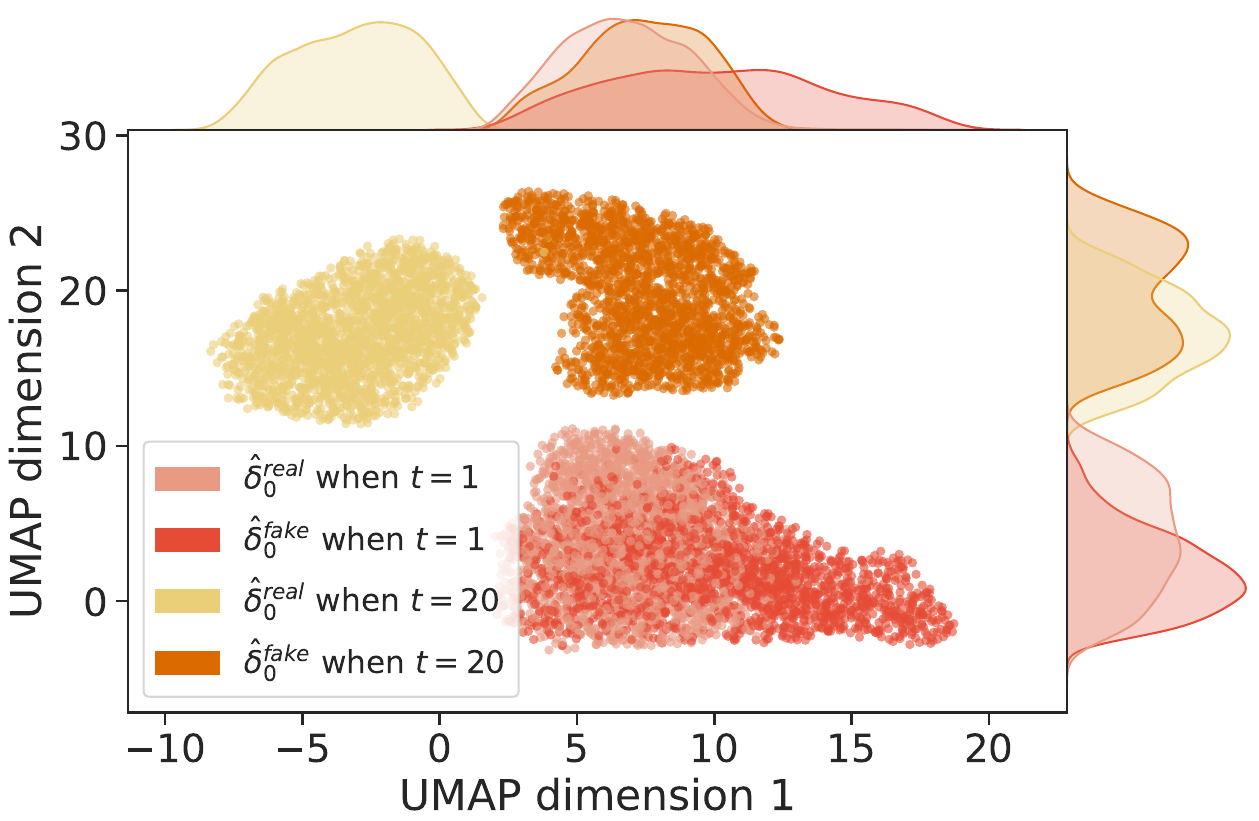}}
\hfil
\subfloat[Our residual bias $\Delta$]{\includegraphics[width=0.8\linewidth]{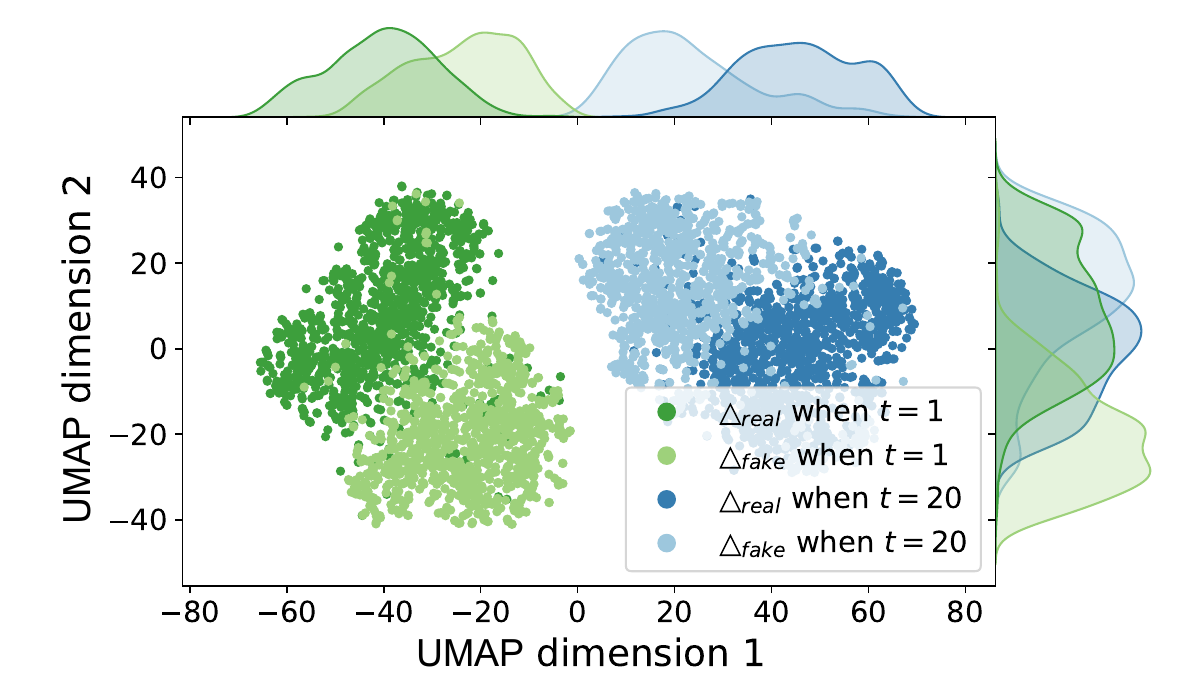}}
\caption{UMAP~\cite{mcinnes2018umap} visualization of real vs. fake features at step 1 and step 20. (a) Existing methods using measured residuals show poor separability at step 1, requiring multiple steps to distinguish real from fake. (b) Our residual bias features exhibit clear separation even at step 1, enabling efficient and effective detection.}
\label{fig:t-sne}
\end{figure}


\textbf{Theoretical Analysis.} Given an input image $x_0^I$, we apply DDIM inversion to project it into a noise vector $x_t^I$, and then perform DDIM sampling to obtain a reconstructed image $x_0^R$. The forward reconstruction process is initialized directly from the inverted noise, i.e., $x_t^R = x_t^I$. We define the theoretical residual at step $t$ as $\delta_t  = x_t^I - x_t^R$. Through recursive expansion, the total residual at step $t = 0$ can be expressed as (derivation details in Appendix:
\begin{equation}
\small
\begin{aligned}
\delta_{0} = &\sum_{i=1}^{t} \mathcal{F}_i ,\\
\text{where } \mathcal{F}_i = &\frac{\left ( \sqrt{1-\bar{\alpha}_{i}}-\sqrt{\alpha_{i}-\bar{\alpha}_{i}} \right ) \cdot  \left( \epsilon_\theta\left(x_{i}^R,i\right)-\epsilon_\theta\left(x_{i-1}^I,i\right)\right)}{\sqrt{\bar{\alpha}_{i}}}.
\end{aligned}
\label{eq:residual_0}
\end{equation}
which depends solely on model noise predictions and pre-defined sampling parameters.
In contrast, the measured residual is defined as the pixel-wise difference between the original and reconstructed images: $\hat{\delta}_0 = x_0^I - x_0^R.$

Existing reconstruction-based methods use measured residuals $\hat{\delta}_0$ directly for classification. However, we argue this quantity is entangled: it contains both (i) reconstruction errors caused by imperfect inversion and sampling, and (ii) real-domain specific clues that exist only in real images. This entanglement necessitates multiple reconstruction steps to sufficiently amplify the differences between real and fake images, leading to a high inference cost (as illustrated in Fig.~\ref{fig:t-sne}(a)).
When using only one reconstruction step, both real and fake images exhibit similar levels of residual noise, making them difficult to distinguish. 

To address this, we propose to decouple the real-domain semantic clues from the reconstruction residual by defining a new discriminative metric, i.e., residual bias
$
\Delta = \left| \hat{\delta}_0 - \delta_0 \right|.
$
This residual bias captures the discrepancy attributable to domain-specific information in real images, which the generative model is unable to reproduce. We visualize the residual bias under different reconstruction steps in Fig.~\ref{fig:t-sne}(b). Notably, residual bias reveals well-separated clusters for real and fake images even when computed with only a single inference step, enabling significantly more efficient detection compared to prior methods that require hundreds of steps.

\textbf{Residual Bias in Two Spaces.} 
In practice, we use a pre-trained G-LDM as the reconstruction model. A key challenge arises: $\delta_0$ and $\hat{\delta}_0$ lie in different spaces, where $\delta_0$ is computed in latent space using predicted noise, while $\hat{\delta}_0$ is measured in RGB space via pixel-wise differences. Therefore, they cannot be directly subtracted to obtain a valid residual bias. 
To resolve this, we compute residual bias in both latent and RGB spaces, by projecting the original and reconstructed images into the same space via the VAE encoder/decoder used in the latent diffusion model. This yields two variants:
\begin{equation}
\small
\begin{aligned}
\Delta_{Latent} = &\left|\mathcal{E}\left(\hat{\delta}_0\right) - \delta_0 \right|= \left| \mathcal{E}(x_0^I) - \mathcal{E}(x_0^R) -  \sum_{i=1}^{t} \mathcal{F}_i\right|,\\
\Delta_{RGB} = & \left|\hat{\delta}_0 - \mathcal{D}(\delta_0)\right| = \left|x_0^I - x_0^R - \mathcal{D}(\delta_0)\right|.
\end{aligned}
\end{equation}
where $\mathcal{E}(\cdot)$ denotes the encoder and $\mathcal{D}(\cdot)$ denotes the decoder of VAE in G-LDM model. These complementary residual bias estimates are then fused as input features to a downstream classifier, as shown in Fig.~\ref{fig:inference}.

\section{Evaluation}
\label{sec:eval}
In this section, we conduct extensive experiments to empirically evaluate the proposed R$^2$BD. Broadly, we intend to answer the following research questions:
\begin{itemize}[leftmargin=*,itemsep=2pt,topsep=0pt,parsep=0pt]
    \item \textbf{RQ1:} How well does R$^2$BD generalize across different generative paradigms, especially when trained on one type (e.g., GAN, pixel diffusion model, or latent diffusion model) and tested on unseen generators?
    \item \textbf{RQ2:} Can R$^2$BD benefit from hybrid training on multiple generator types to enhance its overall detection performance?
    \item \textbf{RQ3:} 
    As two key components of R$^2$BD, how does the rationality of G-LDM and residual bias calculation module?
    \item \textbf{RQ4:} How robust is R$^2$BD to common perturbations in real-world scenarios?
    \item \textbf{RQ5:} How much do different components of R$^2$BD contribute to its performance?
    \item \textbf{RQ6:} What are the limitations of R$^2$BD?
\end{itemize}
Next, we first present experimental settings, followed by answering the above research questions one by one. 


\begin{table*}[t]
\caption{Composition of the evaluated dataset. }
\centering
\setlength\tabcolsep{4pt}
\begin{tabular}{cccccc}
\toprule
Split                                                                                              & Type                                                                  & Method                   & Source             & \# of images            & Resolution \\ \midrule
\multirow{6}{*}{\begin{tabular}[c]{@{}c@{}}Train / In-dataset Test\end{tabular}} & \multirow{2}{*}{Real}                                                 & -                        & CelebA-HQ~\cite{karras2018progressive}          & 10,000/2,000                  & 1024$\times$1024  \\
                                                                                                   &                                                                       & -                        & CASIA-WebFace~\cite{yi2014learning}      & 10,000/2,000                  & 250$\times$250    \\\cmidrule(r){2-6}
                                                                                                   & GAN                                                                   & StyleGAN2~\cite{karras2020analyzing}                & SFHQ~\cite{david_beniaguev_2022_SFHQ}               & 10,000/2,000                  & 1024$\times$1024  \\\cmidrule(r){2-6}

                                                                                                   & \begin{tabular}[c]{@{}c@{}}Pixel Diffusion Model (PixelDM)\end{tabular}                   & ADM~\cite{dhariwal2021diffusion}                      & DiffusionFace~\cite{chen2024diffusionface}      & 10,000/2,000                  & 256$\times$256    \\ \cmidrule(r){2-6}
                   & \multirow{2}{*}{\begin{tabular}[c]{@{}c@{}}Latent Diffusion Model(LatentDM)\end{tabular}} & \multirow{2}{*}{Stable Diffusion V2.1~\cite{rombach2022high}} & DiffusionForensics~\cite{wang2023dire} & 5,000/1,000                   & 768$\times$768    \\
                                                                                                   &                                                                       &                          & SFHQ~\cite{david_beniaguev_2022_SFHQ}               & 5,000/1,000                   & 1024 $\times$1024  \\\midrule
\multirow{14}{*}{\begin{tabular}[c]{@{}c@{}}Cross-dataset Test\end{tabular}}                  & Real                                                                  & -                        & VggFace2~\cite{cao2018vggface2}           & 10,000                  & Unfixed    \\\cmidrule(r){2-6}
                                                                                                   & \multirow{4}{*}{GAN}                                                  & ProGAN~\cite{karras2018progressive}                   & Kaggle\footnotemark[1]             & 2,000        & 128$\times$128    \\
                                                                                                   &                                                                       & StyleGAN3~\cite{karras2021alias}                & Kaggle\footnotemark[1]           & 2,000        & 256$\times$256    \\
                                                                                                   &                                                                       & StarGAN~\cite{choi2018stargan}                  & ArtiFact~\cite{artifact}           & 2,000        & 200$\times$200    \\
                                                                                                   &                                                                       & Projected GAN~\cite{Sauer2021NEURIPS}            & ArtiFact~\cite{artifact}           & 2,000        & 200$\times$200    \\\cmidrule(r){2-6}

                                                                                                   & \multirow{4}{*}{\begin{tabular}[c]{@{}c@{}}Pixel Diffusion Model(PixelDM)\end{tabular}}   & DDPM~\cite{ho2020denoising}                     & DiffusionFace~\cite{chen2024diffusionface}      & 2,000        & 256$\times$256    \\
                                                                                                   &                                                                       & DDIM~\cite{song2021denoising}                     & DiffusionFace~\cite{chen2024diffusionface}      & 2,000        & 256$\times$256    \\
                                                                                                   &                                                                       & PNDM~\cite{liu2022pseudo}                     & DiffusionFace~\cite{chen2024diffusionface}      & 2,000        & 256$\times$256    \\
                                                                                                   &                                                                       & DiffSwap~\cite{zhao2023diffswap}                 & DiffusionFace~\cite{chen2024diffusionface}      & 2,000        & 256$\times$256\\\cmidrule(r){2-6}
                                                                                                                                                                                                      & \multirow{5}{*}{\begin{tabular}[c]{@{}c@{}}Latent Diffusion Model(LatentDM)\end{tabular}}  & LDM~\cite{rombach2022high}                      & DiffusionFace~\cite{chen2024diffusionface}      & 2,000        & 256$\times$256    \\
                                                                                                   &                                                                       & Inpaint~\cite{chen2024diffusionface}                  & DiffusionFace~\cite{chen2024diffusionface}      & 2,000        & 256$\times$256    \\
                                                                                                   &                                                                       & Stable Diffusion V1.5~\cite{rombach2022high}                  & SFHQ~\cite{david_beniaguev_2022_SFHQ}               & 2,000        & 1024$\times$1024  \\
                                                                                                   &                                                                       & Midjourney~\cite{midjouney}               & Self-built         & 2,000        & 1024$\times$1024  \\
                                                                                                   &                                                                       & DALL$\cdot$E 2~\cite{ramesh2022hierarchical}                 & Self-built         & 2,000        & 1024$\times$1024  \\\bottomrule
\end{tabular}
\label{tab:FaceFake-X}
\end{table*}
\footnotetext[1]{\url{https://www.kaggle.com/datasets/mayankjha146025/fake-face-images-generated-from-different-gans}}

\subsection{Setup}

\textbf{Data Preparation.}
To comprehensively evaluate the generalization and robustness of AIGC detectors, we construct an evaluation dataset by aggregating face images from a total of 11 different sources, including 8 publicly available AIGC datasets and 2 commercial T2I services (Midjourney~\cite{midjouney} and DALL$\cdot$E 2~\cite{ramesh2022hierarchical}). These sources span three major families of generation mechanisms, namely GANs, pixel diffusion models (PixelDM), and latent diffusion models (LatentDM), and cover a wide range of typical forgery applications such as full-image generation, face swapping, attribute editing, and image restoration. For commercial models, we curated prompts along nine demographic and semantic dimensions to generate diverse identities and styles via public APIs (details in Appendix). The dataset includes both high-resolution and low-resolution images, enabling evaluation under varied visual quality and real-world deployment conditions. We split the dataset into a training set and two test sets: an in-dataset test set sharing the same generation methods as training, and a cross-dataset test set comprising 13 unseen forgery methods from different distributions, as listed in Tab.~\ref{tab:FaceFake-X}. Real images were also drawn from three distinct datasets to ensure resolution and distribution diversity. Overall, this benchmark allows us to rigorously assess a detector’s cross-method, cross-resolution, and cross-distribution generalization in realistic settings.


\textbf{Implementation Details.}  For the proposed G-LDM, we use the image-text pair generation dataset and the CelebA-HQ real dataset for training, as detailed in Appendix. Initial parameters are taken from Stable Diffusion V2.1 and UNet is fine-tuned. The discriminator in G-LDM adopts the model architecture and gradient penalty loss from WGAN~\cite{gulrajani2017improved}. G-LDM was trained for a total of $10$ epochs with a learning rate of $5 \times 10^{-5}$. Furthermore, in line with previous works~\cite{wang2023dire,yang2024gaussian}, we also use empty text to reconstruct images by the pre-trained G-LDM. To train the two-stream network, we employed ResNet-18~\cite{he2016deep} as the backbone model, training it for $20$ epochs with a learning rate of $1 \times 10^{-4}$.

\begin{table}[th]
\centering
\setlength\tabcolsep{4pt}
\caption{Comparison of detection performance between existing methods and our proposed R$^2$BD under cross-paradigm generalization settings. The "Training dataset" column denotes detectors trained on a single generative paradigm, including GAN (StyleGAN2), PixelDM (ADM), or LatentDM (Stable Diffusion v2.1). The "In-dataset" column reports performance on the same generative method as used in training, while the "Cross-dataset" columns evaluate generalization to 13 unseen generative methods. We report ACC(\%) for both in-dataset and cross-dataset settings. Gray cells highlight within-paradigm generalization across different algorithms of the same family.
}
\begin{tabular}{cccccc}
\toprule
\multirow{3}{*}{Method} & \multirow{3}{*}{\begin{tabular}[c]{@{}c@{}}Training\\ dataset\end{tabular}} & \multicolumn{4}{c}{Test dataset} \\ \cmidrule(r){3-6}
 &  & \multirow{2}{*}{In-dataset} & \multicolumn{3}{c}{Cross-dataset} \\ \cmidrule(r){4-6}
 &  &  & GAN & PixelDM & LatentDM \\ \midrule
\multirow{3}{*}{Xception} & GAN & 99.50 & \cellcolor{gray!25}59.38 & 59.11 & 62.63 \\
 & PixelDM & 99.77 & 60.04 & \cellcolor{gray!25}80.92 & 55.43 \\
 & LatentDM & 99.30 & 55.89 & 55.43 & \cellcolor{gray!25}58.24 \\\midrule
\multirow{3}{*}{Exposing} & GAN & 99.82 & \cellcolor{gray!25}61.07 & 57.02 & 66.70 \\
 & PixelDM & 99.97 & 59.22 & \cellcolor{gray!25}75.78 & 58.34 \\
 & LatentDM & 100.0 & 58.79 & 55.57 & \cellcolor{gray!25}64.25 \\\midrule
\multirow{3}{*}{\begin{tabular}[c]{@{}c@{}}DeepFake-\\ Adapter\end{tabular}} & GAN & 99.70 & \cellcolor{gray!25}66.57 & 55.63 & 73.82 \\
 & PixelDM & 99.18 & 61.09 & \cellcolor{gray!25}61.71 & 67.01 \\
 & LatentDM & 99.95 & 56.89 & 55.56 & \cellcolor{gray!25}64.84 \\ \midrule
\multirow{3}{*}{DIRE} & GAN & 98.30 & \cellcolor{gray!25}65.68 & 62.27 & 79.06 \\
 & PixelDM & 97.76 & 61.17 & \cellcolor{gray!25}75.28 & 77.66 \\
 & LatentDM & 98.84 & 60.72 & 60.72 & \cellcolor{gray!25}75.39 \\\midrule
ZeroFake & - & - & 60.78 & 68.37 & 63.48 \\\midrule
\multirow{3}{*}{\textbf{R$^2$BD}} & GAN & 97.54 & \cellcolor{gray!25}\textbf{67.86} & 63.74 & 81.02 \\
 & PixelDM & 98.10 & 70.88 & \cellcolor{gray!25}\textbf{90.77} & 67.36 \\
 & LatentDM & 98.98 & 58.73 & 57.17 & \cellcolor{gray!25}\textbf{79.66} \\\bottomrule
\end{tabular}
\label{tab:cross-generator}
\end{table}

\begin{table*}[t]
\centering
\setlength\tabcolsep{1.5pt}
\caption{Evaluation of detection performance under hybrid training across multiple generator types. All detectors are trained on a combined dataset containing samples from three generation paradigms (StyleGAN2, ADM, and Stable Diffusion V2.1), and evaluated on both in-distribution (in-dataset) and out-of-distribution (cross-dataset) samples. We also report the ACC(\%) for the GAN, PixelDM, and LatentDM, respectively. The best results are highlighted in bold.}
\begin{tabular}{ccc|ccccc|ccccc|ccc}
\toprule
\multirow{3}{*}{{Type}}                                                               & \multirow{3}{*}{{Method}} & \multirow{3}{*}{{Venues}} & \multicolumn{5}{c|}{{In-dataset}}                                                                                     & \multicolumn{8}{c}{{Cross-dataset}}                                                                                                                                     \\\cline{4-16}
                                                                                    &            &             & \multirow{2}{*}{ACC$\uparrow$} & \multirow{2}{*}{AUROC$\uparrow$} & \multirow{2}{*}{AUPRC$\uparrow$} & \multirow{2}{*}{BDR$\uparrow$} & \multirow{2}{*}{EER$\downarrow$} & \multirow{2}{*}{ACC$\uparrow$} & \multirow{2}{*}{AUROC$\uparrow$} & \multirow{2}{*}{AUPRC$\uparrow$} & \multirow{2}{*}{BDR$\uparrow$} & \multirow{2}{*}{EER$\downarrow$} & \multicolumn{3}{c}{ACC$\uparrow$ for each type}            \\ \cline{14-16}
                                                                                    &           &              &                      &                      &                        &                      &                      &                      &                      &                        &                      &                      & GAN            & PDM       & LDM            \\ \hline
\multirow{3}{*}{\begin{tabular}[c]{@{}c@{}}Feature\\ extraction-based\end{tabular}} & Xception     & ICCV 2019           & 99.11              & 99.93              & 99.96                & 99.30              & 0.90               & 68.02              & 87.41              & 95.21                & 97.51              & 19.37              & 62.72          & 82.02          & 82.02          \\
                                                                                    & Exposing      & AAAI 2024          & \textbf{99.78}     & 99.83    &       99.88
                                                                            & \textbf{99.80}     & {0.61}      & 64.15              & 88.69              & 95.00                & 98.60              & 18.45              & 66.23          & 82.45          & 79.77          
    \\
                                                                                    & DeepFake-Adapter     &   IJCV 2025           & 99.66     & 99.95    &       \textbf{99.96}
                                                                            & {99.52}     & \textbf{0.38}      & 57.78              & 86.83              & 96.20                & \textbf{99.95}              & \textbf{2.47}              & 70.76          & 60.26          & 86.04\\\hline
\multirow{3}{*}{\begin{tabular}[c]{@{}c@{}}Reconstruction-\\ based\end{tabular}}    & DIRE       &   ICCV 2023             & 98.99              & 99.94              & 99.96                & 99.17              & 1.00               & 64.91              & 86.67              & 95.15                & {99.42}     & 21.80              & 66.57          & 78.32          & 85.81          \\
                                                                                    & ZeroFake     &   CCS 2024           & 72.00              & 79.84              & 86.53                & 81.47              & 27.61              & 67.56              & 73.54              & 87.50                & 84.17              & 32.82              & 61.98          & 73.92          & 66.16          \\
                                                                                    & \textbf{R$^2$BD}  & -     & 98.96              & \textbf{99.96}              & \textbf{99.97}                & 98.40              & 0.75               & \textbf{78.35}     & \textbf{92.30}     & \textbf{96.77}       & 96.66              & {15.46}     & \textbf{72.37} & \textbf{88.91} & \textbf{89.37}
                                                                                    \\ \bottomrule
\end{tabular}
\label{tab:result_FaceFake-X}
\end{table*}

\textbf{Evaluation Metrics.}
Following the recommendations of ~\cite{layton2024sok}, we adopt ACC, AUROC, AUPRC, BDR and EER in our experiments to evaluate the detectors.
1) ACC measures the proportion of correctly classified samples among all samples. It reflects the overall correctness of the detector's predictions.
2) AUROC is the area under the curve, used to evaluate the overall performance of a classification model across various decision thresholds. 
3) AUPRC is the area under the precision-recall curve, suitable for handling imbalanced datasets. It measures the trade-off between precision and recall for positive class predictions.
4) BDR is the bayesian detection rate~\cite{layton2024sok}, which is the probability of an anomaly given the base rate, defined as: $BDR = \frac{P(br)\cdot TPR}{P(br)\cdot TPR+(1-P(br))\cdot FPR}$, where TPR is the true positive rate and FPR is the false positive rate. $P(br)$ is the base rate, i.e., the probability of encountering a sample of deepfake media. We use the deepfake/real ratio to calculate the assumed base-rate of incidence for deepfakes. The base-rate setting is listed in Appendix. 
5) EER is the error rate when the false positive rate is equal to the false negative rate. 
For AUROC, AUPRC, and BDR, values closer to 1 indicate better detector performance. The lower the EER, the better the combined identification performance of the model.


\textbf{Comparison Method.} We select three representative methods for feature extraction-based detection and two state-of-the-art methods for reconstruction-based detection. For feature extraction-based methods, we choose Xception~\cite{rossler2019faceforensics++}, one of the most classic neural networks for deepfake detection; Exposing~\cite{ba2024exposing}, which is specifically designed to enhance generalization; and DeepFake-Adapter~\cite{shao2025deepfake}, which adjusts transferable high-level semantics in large pre-trained Vision Transformers (ViTs) to improve generalization for deepfake detection. For reconstruction-based methods, we select DIRE~\cite{wang2023dire} and ZeroFake~\cite{sha2024zerofake}, both of which leverage pre-trained diffusion models for detection.

\subsection{RQ1: Cross-Paradigm Generalization Ability}

To evaluate the cross-generator generalization performance of different detectors, we adopt a setting where models are trained using fake images from only a single generation method, then tested on unseen generation methods of the same family. Specifically, we consider three generator types: GANs, PixelDMs, and LatentDMs. For each type, we use one representative algorithm in training: StyleGAN2 for GANs, ADM for PixelDM, and Stable Diffusion V2.1 for LatentDM. The corresponding real data includes CelebA-HQ and CASIA-WebFace. The in-dataset test set contains samples from the same distribution, while the cross-dataset test set includes images from unseen generative methods. For example, in the GAN setting, the cross-dataset includes real images from VggFace2 and fake images from ProGAN, StyleGAN3, StarGAN, and Projected GAN. In total, we test on 13 diverse fake generation methods, as listed in Tab~\ref{tab:FaceFake-X}.

Tab.~\ref{tab:cross-generator} shows that when detectors are trained on a single generator type, their performance drops significantly when evaluated on fake images from other generator families. For instance, when trained on StyleGAN2, Xception achieves 99.5\% ACC in-dataset, but only 59.38\% when tested on other GANs, and only 59.11\% and 62.63\% on PixelDM-based and LatentDM-based fakes, respectively. Similar trends are observed for diffusion-based training. These results suggest that different generation families leave distinct forensic traces, making it challenging for detectors to generalize across fundamentally different generation processes. This highlights that in AIGC detection, it is insufficient to train on a single generator type when aiming to robustly detect forgeries from a wide range of generative families.

Interestingly, even within the same generator family, detectors often experience a significant decline in performance when encountering unseen algorithms. For example, while DIRE achieves 98.30\% in-dataset ACC when trained on StyleGAN2, its cross-dataset performance on other GANs drops to 65.68\%. Similarly, training on ADM and testing on other PixelDMs (such as DDPM, DDIM, PNDM, and DiffSwap) leads to a noticeable performance drop, with an ACC decrease of 22.48\%. This suggests that generative methods from the same family, though based on similar underlying principles, introduce significantly different artifacts, making generalization non-trivial. Furthermore, we observe that GAN-generated images are generally more difficult to detect than diffusion-generated ones, as shown by consistently lower ACC across most GAN-based rows.

Although R$^2$BD does not uniformly surpass all baselines across every setting, it consistently demonstrates strong and often leading performance within the same generator family. For instance, when trained on a PixelDM (ADM), R$^2$BD achieves 90.77\% ACC on unseen pixel diffusion methods, substantially outperforming other detectors in that category. Similar trends are observed for GANs and LatentDMs, where R$^2$BD maintains state-of-the-art accuracy across intra-family generalization tasks. These results suggest that R$^2$BD is capable of capturing certain generator-invariant features within the same generation paradigm, leading to relatively strong generalization even when trained on limited representative data.

This insight provides important implications for real-world deployment. In practice, it is infeasible to anticipate and train on every specific generation algorithm. However, the majority of fake image generation techniques are built upon a few canonical generative mechanisms, such as GANs, pixel-space diffusion models, and latent-space diffusion models. Instead of exhaustively covering individual algorithms, a more scalable and effective strategy is to focus on representative generators from each family, enabling detectors to learn transferable features that generalize to unseen methods within the same paradigm.

\begin{center}
\fcolorbox{black}{gray!10}{\parbox{.98\linewidth}{\textbf{Summary}: 
Different generative paradigms produce distinct artifacts, and even within the same paradigm, unseen algorithms pose generalization challenges. R$^2$BD mitigates this and shows relatively strong intra-paradigm generalizability. }}
\end{center}

\subsection{RQ2: Generalization Ability under Hybrid Training}

Building upon the cross-paradigm setting in RQ1, we now evaluate whether training on a combination of generator types can further improve the generalization ability of detectors. Specifically, all methods are trained on a mixed dataset that includes fake images from three representative generators: StyleGAN2 (GAN), ADM (pixel diffusion), and Stable Diffusion V2.1 (latent diffusion), and tested on the same 13 unseen generation methods as in the cross-dataset setting of RQ1.

As shown in Tab.~\ref{tab:cross-generator}, training detectors on a combined dataset containing multiple generator types significantly improves their cross-dataset generalization performance. Compared to single-generator training (as in RQ1), all methods show enhanced ability to detect forgeries from previously unseen algorithms. R$^2$BD achieves the highest overall accuracy (78.35\%) across unseen data, along with top AUROC (92.30\%) and AUPRC (96.77\%). These results indicate that exposing models to a variety of generation paradigms during training enables better generalization beyond the training distribution.

Breaking down the performance by generator type, R$^2$BD achieves the highest cross-dataset accuracy on all three generator families: 72.37\% for GANs, 88.91\% for PixelDMs, and 89.37\% for LatentDMs. This demonstrates its balanced ability to handle diverse generation paradigms. In contrast, other detectors often exhibit modality-specific biases: for example, Exposing performs well on PixelDMs but struggles on LatentDMs, while DIRE performs unevenly across GAN and diffusion-based fakes. This highlights R$^2$BD’s capacity to maintain consistent generalization across all types, rather than overfitting to a particular generation mechanism. Furthermore, compared to other reconstruction-based approaches such as DIRE and ZeroFake, R$^2$BD’s advantage lies in its explicit modeling of the generation paradigm. While other reconstruction-based methods such as DIRE and ZeroFake rely primarily on pre-trained diffusion models, R$^2$BD offers a more principled approach by simulating multiple generation paradigms.

\begin{center}
\fcolorbox{black}{gray!10}{\parbox{.98\linewidth}{\textbf{Summary}: 
Hybrid training on diverse generator types significantly improves generalization to unseen forgeries. R$^2$BD achieves consistently strong performance across all generation paradigms.}}
\end{center}

\subsection{RQ3: Rationality Analysis}\label{sec:RQ5}
Recall that the G-LDM and residual bias calculation module are designed to enhance generalization and detection efficiency, respectively. In this section, we analyze the rationality and effectiveness of these two core components in R$^2$BD.

\begin{figure*}[t]
  \centering
  \includegraphics[width=0.9\linewidth]{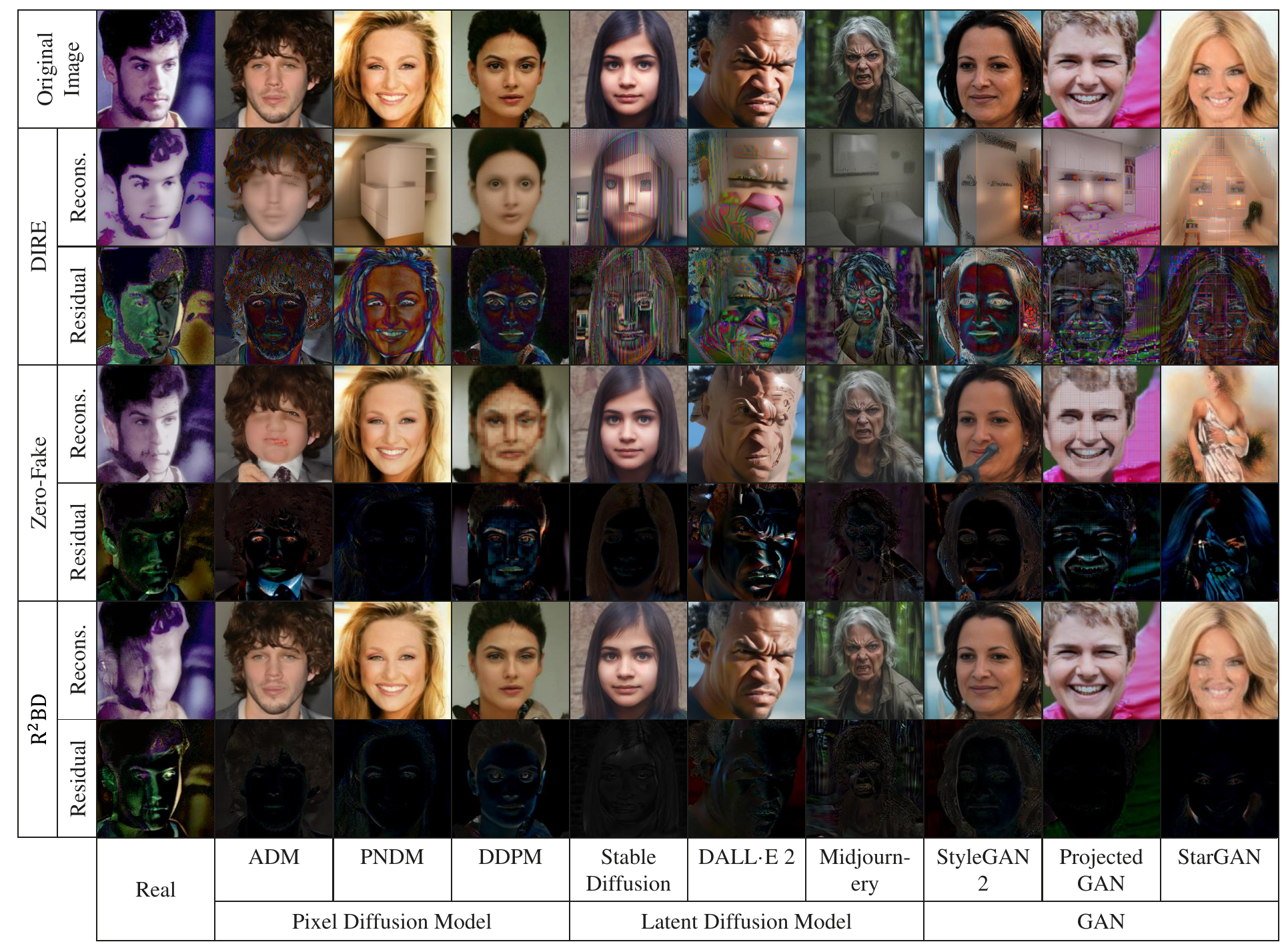}
  \caption{The visualisation results of the proposed R$^2$BD and related works.}
  \label{fig:visualisation}
\end{figure*}

\textbf{Visualisation Analysis.} To evaluate the generalization capability introduced by G-LDM, we visualize reconstructed images and their corresponding residuals across different reconstruction strategies. For fair comparison, we follow the experimental setups from the original DIRE and ZeroFake papers. Specifically, we set $t=50$ for reconstruction using our G-LDM and obtain residuals accordingly. We compare real images with fake images generated by nine forgery methods, covering all three paradigms: PixelDM, LatentDM, and GAN (see Fig.~\ref{fig:visualisation}). The results show that DIRE performs well when reconstructing images generated by PixelDMs (particularly DDPM), because it uses ADM as the reconstruction backbone. However, for GAN- and LatentDM-based forgeries, the reconstruction quality of DIRE is poor, resulting in residuals similar to those of real images, which weakens the separability for downstream detection. ZeroFake, which incorporates Stable Diffusion V1.4 with text conditioning, shows slightly better generalization than DIRE. Nevertheless, its residual maps still fail to clearly distinguish between real and fake when the forgery methods deviate from Stable Diffusion (e.g., StarGAN, Projected GAN, ADM, DDPM). In contrast, G-LDM produces clearly distinguishable residuals across all nine forgery methods, making the real-fake boundary more evident and thereby improving detectability.

\begin{table}[t]
\centering
\scriptsize
\setlength\tabcolsep{1.5pt}
\caption{Runtime results. For Xception and Exposing, the detection process simply feeds the image into the detector for classification. For DIRE, ZeroFake, and our R$^2$BD, the detection process contains the reconstruction inference process and classification. The total reconstruction inference process needs DDIM inversion and reconstruction process.}
\begin{tabular}{cc|ccc|ccc}
\toprule
\multirow{2}{*}{Type}  & \multirow{2}{*}{Method}  & \multicolumn{3}{c|}{Inference Step}& \multicolumn{3}{c}{Detecting Time (s)} \\\cline{3-8}
                &               &     Inverse        & Recons.& Total   & Recons.    & Classify        & Total\\\hline
\multirow{3}{*}{\begin{tabular}[c]{@{}c@{}}Feature\\extraction-based\end{tabular}}                     &Xception                                 & -             & -                 & -& -          & {0.677}        & {0.677}\\
&Exposing                                 & -            & -                 & -& -          & 0.881        & 0.881\\

&DeepFake-Adapter & -            & -                 & -& -          & 0.331        & \textbf{0.331}\\
\hline
\multirow{3}{*}{\begin{tabular}[c]{@{}c@{}}Reconstruction-\\based\end{tabular}}&DIRE                                 & 20             & 20                 & 40& 14.93          & 0.964        & 15.90\\
&ZeroFake                          & 999            & 20                 & 1019& 82.29          & \textbf{0.036}        & 82.33 \\
&\textbf{R$^2$BD}                                & \textbf{1}              & \textbf{1}                  & \textbf{2}&\textbf{0.616}          & {0.090}       & {0.706}  \\ \bottomrule  
\end{tabular}
\label{tab:runtime}
\end{table}

\textbf{Detection Time Analysis.} We further evaluate the detection efficiency of R$^2$BD by comparing its runtime with state-of-the-art methods. Feature extraction-based methods (e.g., Xception, Exposing, DeepFake-Adapter) require only a forward pass through a classifier. In contrast, reconstruction-based detectors involve both image reconstruction and classification, where the reconstruction stage typically includes DDIM inversion and generation steps. As shown in Tab.~\ref{tab:runtime}, R$^2$BD achieves efficient detection, taking only 0.706 seconds per image, which is comparable to feature-based detectors like Xception, Exposing, and DeepFake-Adapter. In contrast, DIRE and ZeroFake are significantly slower, with runtimes approximately 22$\times$ and 116$\times$ longer than R$^2$BD, respectively. This efficiency gain stems from the residual bias calculation module in our framework, which reduces the number of inference steps to just $\frac{1}{20}$ of DIRE and $\frac{2}{1019}$ of ZeroFake.

\begin{center}
\fcolorbox{black}{gray!10}{\parbox{.98\linewidth}{\textbf{Summary}: Our G-LDM enables accurate reconstruction across diverse forgery methods, resulting in clear residual differences between real and fake images. The residual bias module ensures high efficiency, requiring only 0.706 seconds per image for R$^2$BD.}}
\end{center}


\begin{figure*}[t]
  \centering
  \subfloat[JPEG Compression]{
  \includegraphics[width=0.19\linewidth]{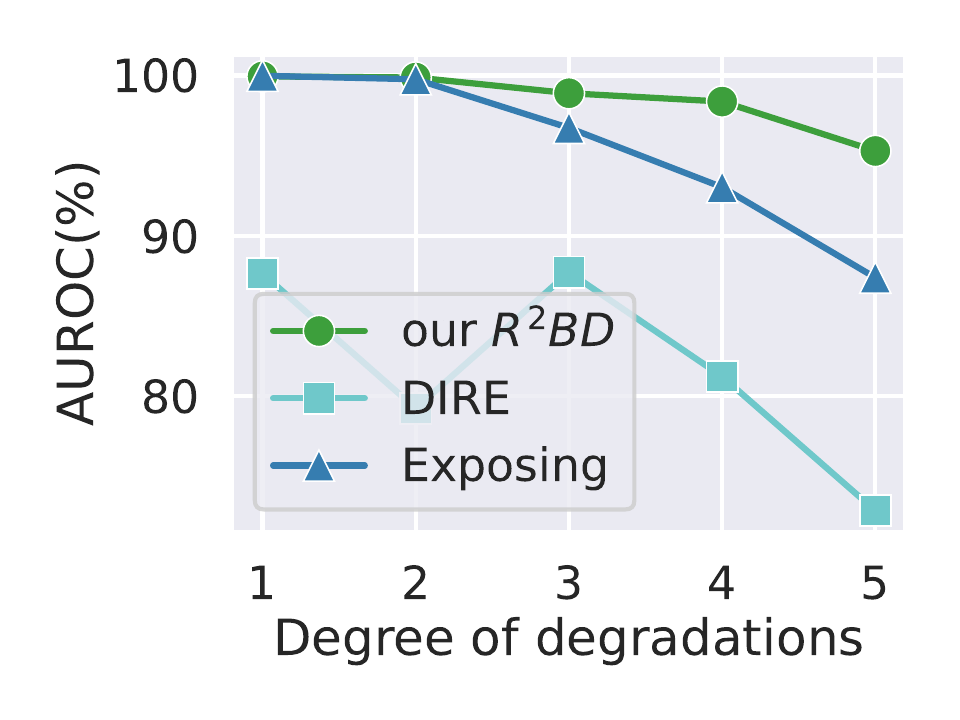}}
  \subfloat[Gaussian Noise]{
  \includegraphics[width=0.19\linewidth]{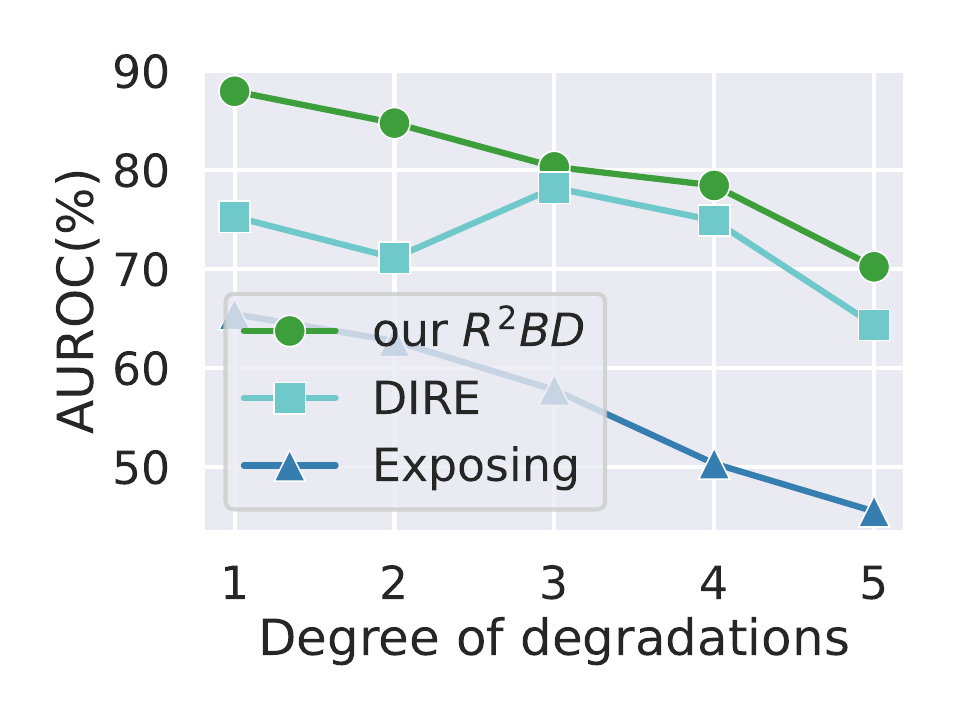}}
  \subfloat[Gaussian Blur]{
  \includegraphics[width=0.19\linewidth]{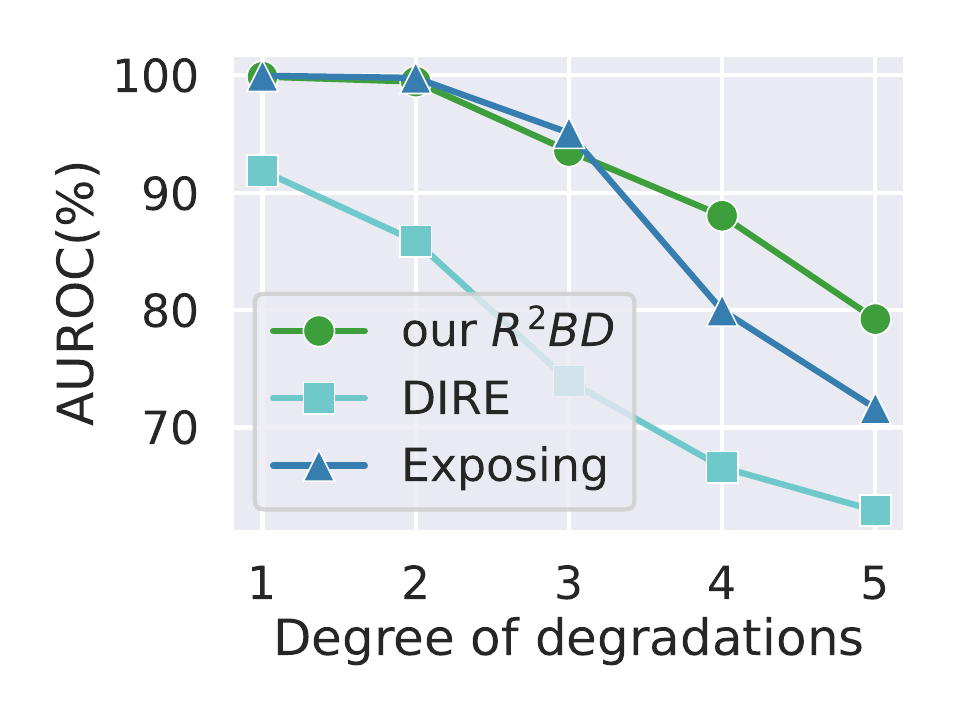}}
  \subfloat[Color Contrast]{
  \includegraphics[width=0.19\linewidth]{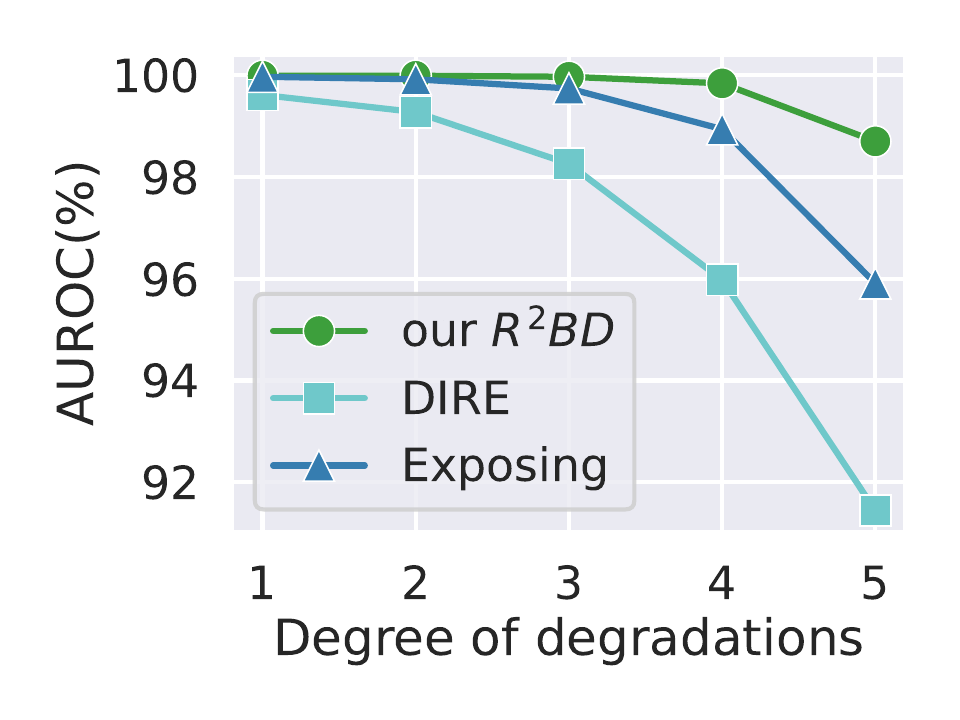}}
  \subfloat[Color Saturation]{
  \includegraphics[width=0.19\linewidth]{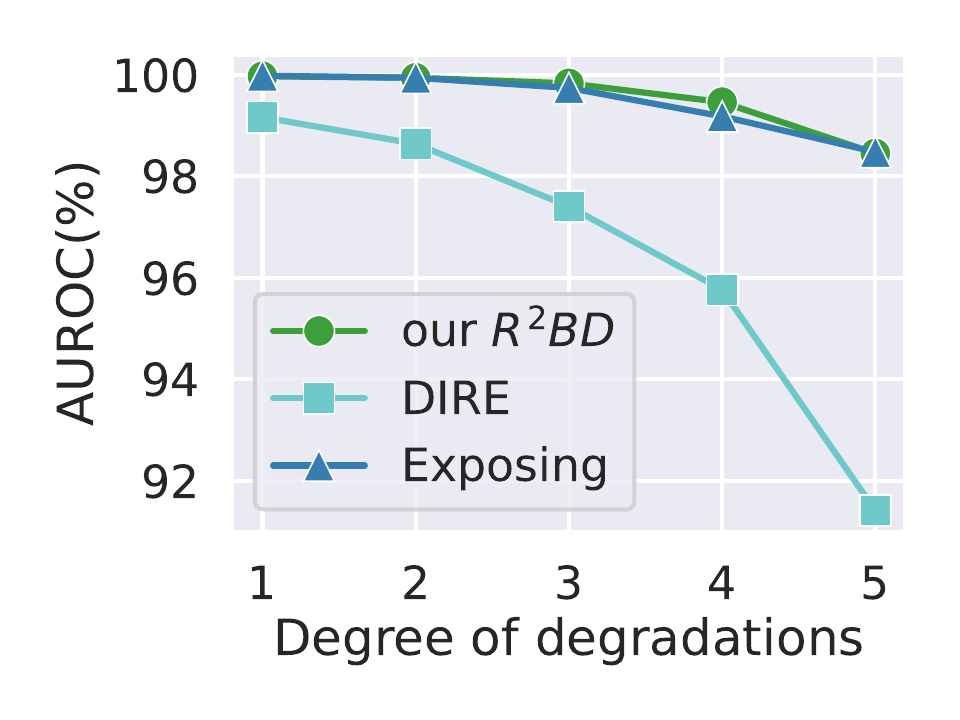}}
  \caption{The AUROC(\%) of robustness evaluation against real-world image degradations.}
  \label{fig:robustness}
\end{figure*}

\begin{table*}[t]
\centering
\setlength\tabcolsep{3pt}
\caption{The ablation study of the proposed G-LDM, Residual Bias Calculation module, Two-stream network for our R$^2$BD. The best results are highlighted in bold.}
\small
\begin{tabular}{ccccc|ccccc|ccccc}
\toprule
\multirow{2}{*}{ID} & \multirow{2}{*}{G-LDM} & \multirow{2}{*}{\begin{tabular}[c]{@{}c@{}}Residual Bias\\ Calculation\end{tabular}} & \multicolumn{2}{c|}{Two-stream} & \multicolumn{5}{c|}{In-dataset} & \multicolumn{5}{c}{Cross-dataset} \\ \cline{4-5}\cline{6-10}\cline{11-15}
 &  &  & RGB & Latent & ACC$\uparrow$ & AUROC$\uparrow$ & AUPRC$\uparrow$ & BDR$\uparrow$ & EER$\downarrow$ & ACC$\uparrow$ & AUROC$\uparrow$ & AUPRC$\uparrow$ & BDR$\uparrow$ & EER$\downarrow$ \\\cline{1-15}
1 & $\times$  & $\surd$  & $\surd$ & $\surd$ & 95.34 & 98.96 & 99.29 & 95.69 & 4.88 & 67.77 & 85.48 & 94.20 & 94.36 & 22.26 \\
2 & $\surd$ &  $\times$ & $\surd$ & -  & 59.85 & 50.05 & 59.81 & 60.00 & 0.88 & 72.22 & 50.63 & 84.23 & 60.00 & 0.51 \\
3 & $\surd$ &  $\surd$ & $\times$ & $\surd$ & 95.38 & 99.02 & 99.32 & 95.48 & 4.60 & 68.65 & 89.15 & 95.32 & 96.20 & 18.61 \\
4 &$\surd$ &  $\surd$ &  $\surd$&$\times$ & 48.51 & 64.25 & 79.83 & 99.88 & 40.75 & 52.94 & 75.07 & 90.10 & 97.58 & 32.23 \\
5 &$\surd$ &  $\surd$ &  $\surd$&$\surd$ & 98.96 & 99.96 & 99.97 & 98.40 & 0.75 & 78.35 & 92.30 & 96.77 & 96.66 & 15.46\\\bottomrule
\end{tabular}
\label{tab:ablation}
\end{table*}

\subsection{RQ4: Robustness to Image Degradations}

In this section, we evaluate the robustness of our R$^2$BD against various image degradations, which is common in the spread of deepfake images and videos. Following the image degradation strategies in ~\cite{jiang2020deeperforensics,dong2022protecting}, we employ five real-world perturbations including jpeg compression, gaussian noise, gaussian blur, color contrast, and color saturation. We set each perturbation to five intensity levels, with higher levels increasing the intensity of the perturbation. All detectors are trained on the same multi-paradigm training set introduced in earlier sections, and evaluated on perturbed versions of the in-distribution test set. The results are shown in Fig.~\ref{fig:robustness}.

First, we can see that the detection performance of the detectors decreases to varying degrees as the intensity of the perturbation increases. For Xception, the input is the original image, and increasing the perturbations weakens the forgery clues on the image, so the difficulty of detection gradually increases. For DIRE, the input is the residual between the reconstructed image and the original image. Most of the pixel values in the residual image are close to 0, so perturbations have a greater impact on it and have the worst robustness. In particular, DIRE has abnormal performance for gaussian noise because the diffusion model used generates images by denoising.

Secondly, our R$^2$BD shows better robustness than the baseline methods. Specifically, R$^2$BD is insensitive to perturbations of jpeg compression, color contrast, and color saturation. R$^2$BD's AUROC decreases by 4.65\%, 1.29\%, and 1.52\%, while the AUROC of Exposing decreases by 12.59\%, 4.06\%, and 1.5\%, respectively. For gaussian noise and gaussian blur, R$^2$BD is relatively more stable. At the fifth level of the most blurry, the AUROC of R$^2$BD is 7.59\% and 16.31\% higher than that of Exposing and DIRE, respectively. At the fifth level of the strongest noise, the AUROC of Exposing and DIRE is 24.72\% and 5.88\% lower than that of R$^2$BD, respectively.

\begin{center}
\fcolorbox{black}{gray!10}{\parbox{.98\linewidth}{\textbf{Summary}: 
R$^2$BD exhibits outstanding robustness to various real-world image perturbations.}}
\end{center}

\subsection{RQ5: Ablation Study}

In this section, we perform ablation evaluation on each component of R$^2$BD, i.e., G-LDM, residual bias calculation module, and two-stream network, as shown in Tab.~\ref{tab:ablation}.

For the control experiment of G-LDM (ID 1 vs. ID 5), we train the detector by reconstructing the images using Stable Diffusion 2.1 as our control group, which corresponds to using the proposed pre-trained G-LDM as the reconstruction model. We can see that the main contribution of G-LDM is to improve the generalization. When using G-LDM, the AUROC is improved by 6.83\% and the EER is decreased by 6.8\% under the cross-dataset setting.


For the control experiment comparing the residual bias calculation module (ID 2 vs. ID 5), we set $t=1$ in the DDIM inversion and reconstruction process to obtain measured residuals, as commonly used in prior works. In ID 2, these residuals were used as input for detection. However, the model performs poorly, with in-dataset AUROC of 50.05\% and ACC of only 59.85\%, which is close to random guessing. This further indicates that measured residuals require accumulation over multiple steps to reveal meaningful real–fake differences. In contrast, ID 5 employs our proposed residual bias, computed at $t=1$, which yields significantly better results: AUROC improves to 99.96\% in-dataset and 92.30\% cross-dataset, with consistent gains across all metrics. These results demonstrate that residual bias enables effective detection even with a single reconstruction step, outperforming traditional multi-step residual accumulation.

For the controlled experiments of the two-stream detector, we use only the latent residual bias (ID 3) or only the RGB residual bias (ID 4). The results show that neither stream alone is sufficient, but their combination captures complementary cues essential for accurate and generalizable detection.


\begin{center}
\vspace{-3mm}
\fcolorbox{black}{gray!10}{\parbox{.98\linewidth}{\textbf{Summary}: Each component of the R$^2$BD contributes uniquely to a different aspect of performance, and their integration achieves optimal overall performance.}}
\vspace{-3mm}
\end{center}

\subsection{RQ6: Limitation}
Our R$^2$BD framework determines whether an image is real or fake by reconstructing the image and calculating the reconstruction bias, making it applicable to various fake image detection scenarios, including artistic creation, beyond just facial data. However, since the proposed G-LDM is a variant of LDM, it incurs higher computational costs during training and detection. In future work, we aim to address this limitation by compressing the G-LDM model or exploring alternative reconstruction models.

\section{Conclusion}
\label{sec:conclusion}


In this paper, we present R$^2$BD, a novel and efficient framework for fake image detection that primarily addresses the challenge of high computational cost while also improving generalization. The proposed framework comprises three main components: (1) G-LDM, a paradigm-aware reconstruction model that unifies the generation behaviors of both GANs and diffusion models; (2) a residual bias calculation module that efficiently computes the discrepancy between theoretical and measured residuals at a single inference step; and (3) a lightweight two-stream detector that jointly analyzes residual bias in both the RGB and latent spaces for robust classification. Extensive experiments demonstrate that R$^2$BD is highly efficient, requiring only 0.706 seconds per image for the full detection pipeline, and significantly outperforming existing reconstruction-based methods in terms of runtime. At the same time, R$^2$BD achieves state-of-the-art performance across a wide range of generative methods, including GANs, pixel-space diffusion models, and latent diffusion models, confirming its strong cross-paradigm generalization ability.

\bibliographystyle{plain}
\bibliography{bib}

\appendix
\section{Appendix}
\label{sec:appendix}

\subsection{Theoretical Residual}
\label{sec:derivation}
In this section, we provide the derivation process of the theoretical residual.

According to \cite{song2021denoising}, the deterministic DDIM process is formulated as:
\begin{equation}
\small
\begin{aligned}
z_t = \sqrt{\alpha_{t}}\left ( z_{t-1}- \sqrt{1-\bar{\alpha}_{t-1}} \epsilon_\theta\left(z_{t},t\right)\right ) +\sqrt{1-\bar{\alpha}_{t} } \epsilon_\theta\left(z_{t},t\right),
\end{aligned}
\label{eq:append_ddim}
\end{equation}
where $z_t$ is the latent representation of the image x at step $t$, and $\epsilon_\theta$ is the model prediction noise. $\alpha_{t}\in (0,1]$ is the noise attenuation rate at step $t$, which in general decreases monotonically with $t$, and $\bar{\alpha}_t = \prod_{i=1}^{t} \alpha_i$.  The DDIM inversion can be formulated as:
\begin{equation}
\small
\begin{aligned}
z_t^I = \sqrt{\alpha_{t}}\left ( z_{t-1}^I- \sqrt{1-\bar{\alpha}_{t-1}} \epsilon_\theta\left(z_{t-1}^I,t\right)\right ) +\sqrt{1-\bar{\alpha}_{t}} \epsilon_\theta\left(z_{t-1}^I,t\right).
\end{aligned}
\label{eq:append_inverse}
\end{equation}

Combined with the above Eq.~\ref{eq:append_ddim} and Eq.~\ref{eq:append_inverse}, the reconstruction process is first inversion process i.e. $ z_0^I \to z_t^I $ and then sample process i.e. $ z_T^R \to z_0^R $.
We define the theoretical residual is the difference between original image and reconstructed image, formulated as:
\begin{equation}
\scriptsize
    \begin{aligned}
\delta _t = & z_t^I -  z_t^R\\
=& \left \{ \sqrt{\alpha_{t}}\left ( z_{t-1}^I- \sqrt{1-\bar{\alpha}_{t-1}} \epsilon_\theta\left(z_{t-1}^I,t\right)\right ) +\sqrt{1-\bar{\alpha}_{t} } \epsilon_\theta\left(z_{t-1}^I,t\right) \right \} \\
&-  \left \{ \sqrt{\alpha_{t}}\left ( z_{t-1}^R- \sqrt{1-\bar{\alpha}_{t-1}} \epsilon_\theta\left(z_{t}^R,t\right)\right ) +\sqrt{1-\bar{\alpha}_{t} } \epsilon_\theta\left(z_{t}^R,t\right) \right \}\\
= & \sqrt{\alpha_{t}}\left (  z_{t-1}^I-z_{t-1}^R\right ) +\left ( \sqrt{1-\bar{\alpha_{t}} }-\sqrt{\alpha_{t}-\bar{\alpha_{t}}} \right ) \cdot  \left (  \epsilon_\theta\left(z_{t-1}^I,t\right)   -\right.  \\
& \left. \epsilon_\theta\left(z_{t}^R,t\right) \right )\\
= & \sqrt{\alpha_{t}}\delta _{t-1}+\left ( \sqrt{1-\bar{\alpha_{t}} }-\sqrt{\alpha_{t}-\bar{\alpha_{t}}} \right ) \cdot  \left (  \epsilon_\theta\left(z_{t-1}^I,t\right)  -\epsilon_\theta\left(z_{t}^R,t\right) \right ).  
\end{aligned}
\label{eq:appendix}
\end{equation}
Therefore, $\delta _{t-1}$ can be further formulated as:
\begin{equation}
\scriptsize
\begin{aligned}
\delta _{t-1} =  \frac{\delta _{t} + \left ( \sqrt{1-\bar{\alpha}_{t}}-\sqrt{\alpha_{t}-\bar{\alpha}_{t}} \right ) \cdot  \left ( \epsilon_\theta\left(z_{t}^R,t\right) - \epsilon_\theta\left(z_{t-1}^I,t\right)   \right )}{\sqrt{\alpha_{t}}}.
\end{aligned}
\label{eq:append_delta _t-1}
\end{equation}
From Eq.~\ref{eq:append_delta _t-1},  we can see that the theoretical residuals gradually accumulate with $t$ and reach a maximum value at $t = 0$. 

We can conduct a case study when $t=3$. According Eq.~\ref{eq:append_delta _t-1}, we can get:
\begin{equation}
\scriptsize
\begin{aligned}
\delta _{2} = & \frac{\delta _{3} + \left ( \sqrt{1-\bar{\alpha}_{3}}-\sqrt{\alpha_{3}-\bar{\alpha}_{3}} \right ) \cdot  \left ( \epsilon_\theta\left(z_{3}^R,3\right) - \epsilon_\theta\left(z_{2}^I,3\right)   \right )}{\sqrt{\alpha_{3}}}.
\end{aligned}
\label{eq:delta_2}
\end{equation}
We can bring Eq.~\ref{eq:delta_2} into $\delta _{1}$ to get:
\begin{equation}
\scriptsize
\begin{aligned}
\delta _{1} = & \frac{\delta _{2} + \left ( \sqrt{1-\bar{\alpha}_{2}}-\sqrt{\alpha_{2}-\bar{\alpha}_{2}} \right ) \cdot  \left ( \epsilon_\theta\left(z_{2}^R,2\right) - \epsilon_\theta\left(z_{1}^I,2\right)   \right )}{\sqrt{\alpha_{2}}}\\
= &\frac{\delta _{3} + \left ( \sqrt{1-\bar{\alpha}_{3}}-\sqrt{\alpha_{3}-\bar{\alpha}_{3}} \right ) \cdot  \left ( \epsilon_\theta\left(z_{3}^R,3\right) - \epsilon_\theta\left(z_{2}^I,3\right)   \right )}{\sqrt{\alpha_{3}\cdot \alpha_{2}}}+\\
&\frac{\left ( \sqrt{1-\bar{\alpha}_{2}}-\sqrt{\alpha_{2}-\bar{\alpha}_{2}} \right ) \cdot  \left ( \epsilon_\theta\left(z_{2}^R,2\right) - \epsilon_\theta\left(z_{1}^I,2\right)   \right )}{\sqrt{\alpha_{2}}}.
\end{aligned}
\label{eq:delta_1}
\end{equation}
The same with $\delta _{0}$:
\begin{equation}
\scriptsize
\begin{aligned}
\delta _{0} = & \frac{\delta _{1} + \left ( \sqrt{1-\bar{\alpha}_{1}}-\sqrt{\alpha_{1}-\bar{\alpha}_{1}} \right ) \cdot  \left ( \epsilon_\theta\left(z_{1}^R,1\right) - \epsilon_\theta\left(z_{0}^I,1\right)   \right )}{\sqrt{\alpha_{1}}}\\
=&\frac{\delta _{3} + \left ( \sqrt{1-\bar{\alpha}_{3}}-\sqrt{\alpha_{3}-\bar{\alpha}_{3}} \right ) \cdot  \left ( \epsilon_\theta\left(z_{3}^R,3\right) - \epsilon_\theta\left(z_{2}^I,3\right)   \right )}{\sqrt{\alpha_{3}\cdot \alpha_{2}\cdot \alpha_{1}}}+\\
&\frac{\left ( \sqrt{1-\bar{\alpha}_{2}}-\sqrt{\alpha_{2}-\bar{\alpha}_{2}} \right ) \cdot  \left ( \epsilon_\theta\left(z_{2}^R,2\right) - \epsilon_\theta\left(z_{1}^I,2\right)   \right )}{\sqrt{\alpha_{2}\cdot \alpha_{1}}}+\\
&\frac{\left ( \sqrt{1-\bar{\alpha}_{1}}-\sqrt{\alpha_{1}-\bar{\alpha}_{1}} \right ) \cdot  \left ( \epsilon_\theta\left(z_{1}^R,1\right) - \epsilon_\theta\left(z_{0}^I,1\right)   \right )}{\sqrt{\alpha_{1}}}\\
=&\frac{\delta _{3}}{\sqrt{\bar{\alpha} _{3}} }  + \sum_{i=1}^{3} \frac{\left ( \sqrt{1-\bar{\alpha_{i}} }-\sqrt{\alpha_{i}-\bar{\alpha_{i}}} \right ) \cdot  \left ( \epsilon_\theta\left(z_{i}^R,i\right) - \epsilon_\theta\left(z_{i-1}^I,i\right)   \right )}{\sqrt{\bar{\alpha} _{i}}}.
\end{aligned}
\end{equation}

Therefore, we can use mathematical induction to derive the $\delta _0$ at the $t$ steps expressed as:
\begin{equation}
\scriptsize
\begin{aligned}
\delta _{0} = & \frac{\delta _{t}}{\sqrt{\bar{\alpha} _{t}} }  + \sum_{i=1}^{t} \frac{\left ( \sqrt{1-\bar{\alpha_{i}} }-\sqrt{\alpha_{i}-\bar{\alpha_{i}}} \right ) \cdot  \left ( \epsilon_\theta\left(z_{i}^R,i\right) - \epsilon_\theta\left(z_{i-1}^I,i\right)   \right )}{\sqrt{\bar{\alpha} _{i}}}.
\end{aligned}
\label{eq:append_delta_0}
\end{equation}
Since $ z_t^I = z_t^R $ at step $t$, then $\delta _t=0$. Thus we can obtain a representation of the final theoretical residuals:
\begin{equation}
\scriptsize
\begin{aligned}
\delta _{0} = &  \sum_{i=1}^{t} \frac{\left ( \sqrt{1-\bar{\alpha_{i}} }-\sqrt{\alpha_{i}-\bar{\alpha_{i}}} \right ) \cdot  \left ( \epsilon_\theta\left(z_{i}^R,i\right) - \epsilon_\theta\left(z_{i-1}^I,i\right)   \right )}{\sqrt{\bar{\alpha} _{i}}}.
\end{aligned}
\label{eq:append_delta_0}
\end{equation}

\subsection{Evaluated Dataset}
\label{sec:appendix_dataset}

\begin{table*}[t]
\centering
\caption{Characteristics of prompts and corresponding options.}
\begin{tabular}{ll}
\toprule
Characteristic & Option                                                                                \\\midrule
Age            & 8, 18, 28, 38, 48, 58, 68, 78, 88                                                     \\
Gender         & Woman (Girl), Man (Boy)                                                                            \\
Human race     & Asian, Black, White                                                                   \\
Emotion        & Happy, Surprised, Sad, Disgusted, Angry, Neutral, Scared                                      \\
Hair color     & White, Black, Red, Brown, Gray, Blond                                                        \\
Background     & Mountain, Forest, Beach, Indoor, Nature, Street, Desert, Sky, City, Space, Underwater \\
Camera type    & polaroid, Leica, Nikon, Hasselblad, GoPro, Kodachrome, Fujifilm, Sony, Canon, Ricoh   \\
Lens type      & 8mm, 28mm, 98mm,148mm, Fisheye Lens                                                   \\
Aperture       & f/1.2, f/1.4, f/2.8, f/4, f/5.6, f/8, f/11, f/16   \\\bottomrule                          
\end{tabular}
\label{tab:prompts}
\end{table*}

\begin{figure*}[t]
  \centering
  \includegraphics[width=\linewidth]{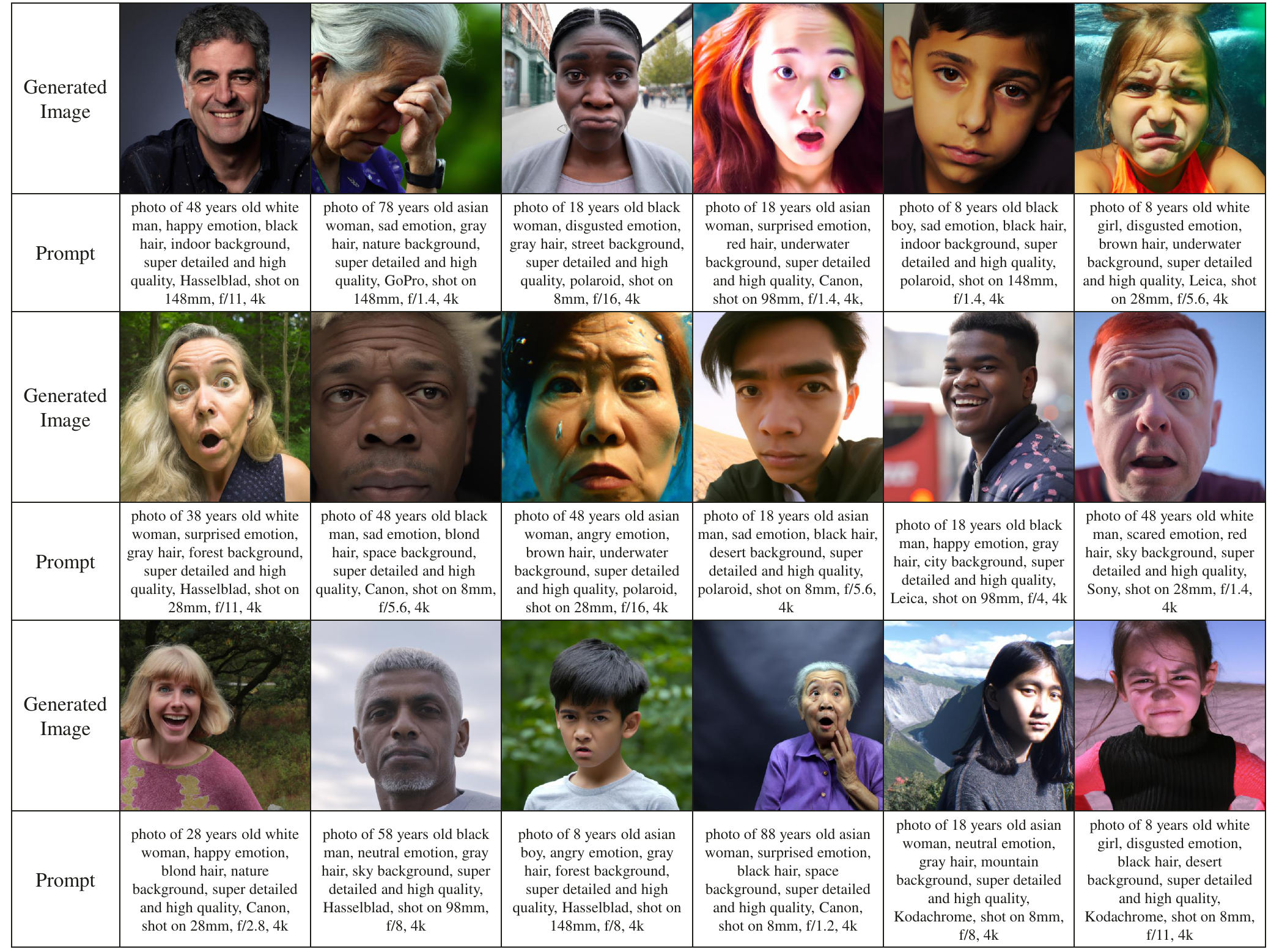}
  \caption{Samples generated by DALL$\cdot$E 2 in the evaluated dataset.}
  \label{fig:dalle2}
\end{figure*}

\begin{figure*}[t]
  \centering
  \includegraphics[width=\linewidth]{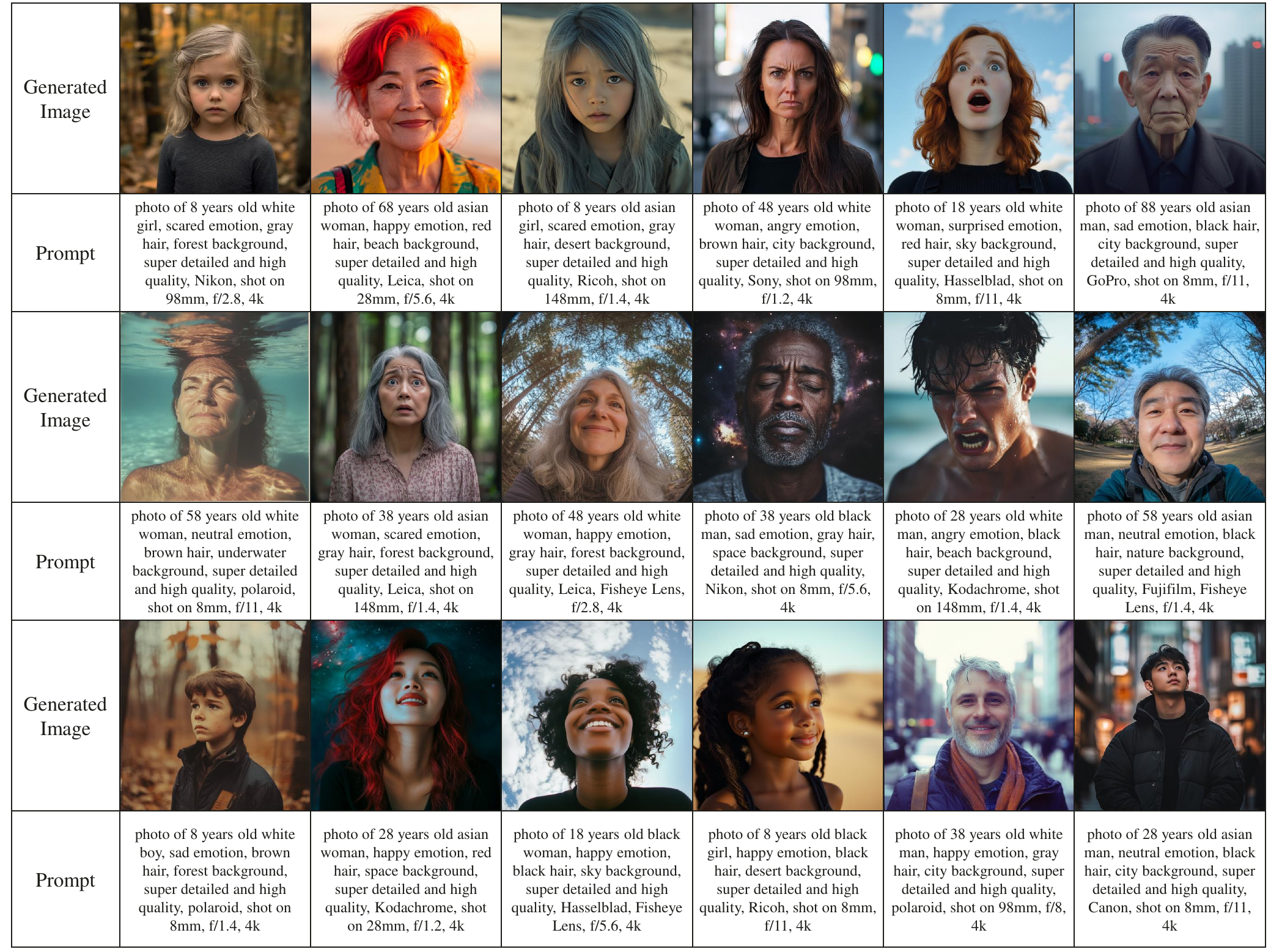}
  \caption{Samples generated by Midjourney in the evaluated dataset.}
  \label{fig:midjourney}
\end{figure*}

Except for the images generated by Midjourney and DALL$\cdot$E 2, the evaluated dataset of the real images and generated images is randomly sampled from public datasets, as detailed composition shown in the main paper.
In addition, we design prompts to generate 2,000 images using Midjourney and DALL$\cdot$E 2 services, respectively. Specifically, we design nine characteristics, as listed in Tab.~\ref{tab:prompts}. We then generated prompts by randomly selecting an option from the corresponding characteristic. We also added the prefix ("photo of") and suffix ("super detailed and high quality, 4k") of the prompt to improve the quality of the generated images. In addition, we randomly sampled some images to show the prompt and the corresponding generated images, as shown in the Fig.~\ref{fig:dalle2} generated by DALL$\cdot$E 2 and the Fig.~\ref{fig:midjourney} generated by Midjourney.

\subsection{Evaluation}
\label{sec:appendix_evaluation}

\paragraph{Data Preprocess}
The training dataset for the G-LDM model comprises both generative and real images. For the generative subset, we collect 200,000 images from SFHQ~\cite{david_beniaguev_2022_SFHQ} and DiffusionFace~\cite{chen2024diffusionface}, which are generated by StyleGAN2, Stable Diffusion V2.1, and ADM, respectively. We then use Llava V1.5~\cite{liu2023llava} to generate descriptive prompts for these images, forming image-text pairs that are used to train the UNet component of G-LDM. For the real subset, we randomly sample 10,000 images from the CelebA-HQ dataset. These real samples, together with the generated images, are used to jointly train the discriminator in G-LDM.

\paragraph{Comparison Method}
1) Xception~\cite{rossler2019faceforensics++}: Xception is the first deep neural network applied to deepfake detection. Due to its outstanding detection performance,  Xception has become a benchmark model in the deepfake detection field. Following the experimental setup in the original paper, we train it for 10 epochs.  
2) Exposing~\cite{ba2024exposing}: Based on the information bottleneck theory, Exposing derives two mutual information loss functions for training the detector, enabling it to capture broader forgery clues. It achieves state-of-the-art performance on FF++, Celeb-DF V2, and DFDC. We also train it for 10 epochs.  
3) DeepFake-Adapter~\cite{shao2025deepfake}: A parameter-efficient method that adapts pre-trained ViTs with lightweight adapters to combine high-level semantics and local forgery cues, achieving strong cross-dataset generalization. We also train it for 10 epochs.
4) DIRE~\cite{wang2023dire}: DIRE uses a pre-trained ADM to reconstruct images and calculates the differences between the reconstructed and original images to train a classifier. We train it for 30 epochs.  
5) ZeroFake~\cite{sha2024zerofake}: ZeroFake employs Stable Diffusion V2.1 to reconstruct images guided by adversarial prompts and uses the similarity between the reconstructed and original images to detect the authenticity of an image. Since ZeroFake requires no training, we use the threshold (0.78) given in the original paper as the similarity score comparison.

\paragraph{Bayesian Detection Rate} The Bayesian Detection Rate (BDR) is proposed by ~\cite{layton2024sok}, which is a measure of the probability that an anomaly is correctly identified, accounting for the base rate of deepfake occurrences. We follow the settings of the base rate in ~\cite{layton2024sok}. Specifically, we set $P(br)=0.6$ and $P(br)=0.722$ on the in-dataset and cross-dataset set of the evaluated dataset, respectively.


 





\end{document}